\documentclass[pdflatex,sn-mathphys-num]{sn-jnl}

\usepackage{graphicx}%
\usepackage{multirow}%
\usepackage{amsmath,amssymb,amsfonts}%
\usepackage{amsthm}%
\usepackage{mathrsfs}%
\usepackage[title]{appendix}%
\usepackage{xcolor}%
\usepackage{textcomp}%
\usepackage{manyfoot}%
\usepackage{booktabs}%
\usepackage{algorithm}%
\usepackage{algorithmicx}%
\usepackage{algpseudocode}%
\usepackage{listings}%

\theoremstyle{thmstyleone}%
\theoremstyle{thmstyletwo}%

\theoremstyle{thmstylethree}%

\begin{document}

\title[Weighted Wasserstein Barycenter of Gaussian Processes for exotic Bayesian Optimization tasks]{Weighted Wasserstein Barycenter of Gaussian Processes for exotic Bayesian Optimization tasks}


\author*[1]{\fnm{Antonio} \sur{Candelieri}}\email{antonio.candelieri@unimib.it}

\author[2]{\fnm{Francesco} \sur{Archetti}}\email{francesco.archetti@unimib.it}
\equalcont{These authors contributed equally to this work.}

\affil*[1]{\orgdiv{Department of Economics Management and Statistics}, \orgname{University of Milano-Bicocca}, \orgaddress{\street{Via Bicocca degli Arcimboldi, 8}, \city{Milan}, \postcode{20126}, \country{Italy}}}

\affil[2]{\orgdiv{Department of Computer Science Systems and Communication}, \orgname{University of Milano-Bicocca}, \orgaddress{\street{Viale Sarca, 336}, \city{Milan}, \postcode{20126}, \country{Italy}}}


\abstract{Exploiting the analogy between Gaussian Distributions and Gaussian Processes' posterior, we present how the weighted Wasserstein Barycenter of Gaussian Processes (W2BGP) can be used to unify, under a common framework, different exotic Bayesian Optimization (BO) tasks. Specifically, collaborative/federated BO, (synchronous) batch BO, and multi-fidelity BO are considered in this paper. Our empirical analysis proves that each one of these tasks requires just an appropriate weighting schema for the W2BGP, while the entire framework remains untouched. Moreover, we demonstrate that the most well-known BO acquisition functions can be easily re-interpreted under the proposed framework and also enable a more computationally efficient way to deal with the computation of the Wasserstein Barycenter, compared with state-of-the-art methods from the Machine Learning literature. Finally, research perspectives branching from the proposed approach are presented.}




\maketitle


\section{Introduction}\label{sec1}
Wasserstein distance \cite{ollivier2014optimal,santambrogio2015optimal,peyre2019computational} has recently and increasingly gained importance in Statistics and Machine Learning. Contrary to statistical divergences, such as Kullback-Liebler, Jensen Shannon, and $\chi^2$, the Wasserstein distance can compare two probability measures while satisfying, under certain assumptions, all the properties of a distance. Moreover, it can also compare probability measures of different types -- two continuous, two discrete, as well as one continuous and one discrete -- while statistical divergences only apply to distributions of the same types. These are the main motivations behind the recent interest in this distance.

Given a set of $M$ probability measures, a first application involving the Wasserstein distance is the computation of their Wasserstein Barycenter (WB), also called the Frechét mean. This is a variational problem that aims to find a probability measure that minimizes the sum of its Wasserstein distances from the $M$ given probability measures. The WB shows many interesting properties, such as \textit{shape preservation} and \textit{geometry awareness}, which are the basis of domain adaptation in Machine Learning \cite{montesuma2024multi} and Generative AI \cite{cheng2024convergence,gao2025wasserstein}.\\

Although its growing relevance in the ML community \cite{arjovsky2017wasserstein,khamis2024scalable,montesuma2024recent}, the use of the Wasserstein distance and the WB in Bayesian Optimization (BO) is still in an embryonic stage. 
A first example is SCoreBO \cite{hvarfner2023self}, which extends Statistical distance-based Active Learning (SAL) to perform BO and Active Learning simultaneously, where Wasserstein distance is one of the distances considered for the SAL step. Other examples are from research work of the authors of this article, specifically for the \textit{gray box} BO \cite{candelieri2022bayesian,candelieri2023wasserstein,candelieri2025bayesian}, where the \textit{gray information} is represented in terms of probability measures and compared via the Wasserstein distance.
Recently, the WB of Gaussian Process (GP) regression models has been proposed to improve the ability of BO to converge to the global optimum by avoiding Maximum Likelihood Estimation to tune the hyperparameters of the GP's kernel \cite{candelieri2025wasserstein}.

On the other hand, there are \textit{exotic} BO tasks that require the combination of different GP models as a core operation, such as Transfer Learning for BO \cite{feurer2018practical,tighineanu2022transfer}, as well as scalable and sparse GP based BO \cite{mcintire2016sparse,wei2024scalable}. Other exotic BO tasks are collaborative/federated BO, batch BO, and multifidelity BO, which are specifically considered in this paper and discussed in related works.

Finally, it is important to clarify that using the WB to combine different GP models is not new: \cite{mallasto2017learning} has already proposed an approach to achieve this goal, but specifically in the learning setting. Their method requires solving an optimization problem over kernel matrices, via a fixed-point method, which could result in a computationally expensive task. Although this cost could be considered reasonable in a learning setting, because it is performed just once or a few times, it can easily become prohibitive in BO, where a GP must be fitted at every iteration.

\newpage

\bmhead{Contributions} The main contributions of this paper can be summarized in the following five points:
\begin{itemize}
    \item presenting a mechanism for computing the WB of GPs more efficiently than \cite{mallasto2017learning}, by exploiting the analogy between univariate Gaussian Distributions (GDs) and the GPs' posterior;
    \item proving how the most widely used acquisition functions can be formulated and interpreted when the WB of GPs is used;
    \item motivating how the \textit{weighted} version of the WB of  GPs allows us to unify, under a unique framework, some of the most well-known exotic BO tasks, specifically collaborative/federated BO, batch BO, and multifidelity BO;
    \item providing an empirical analysis of the impact of different weighting schemes in the \textit{weighted} WB of GPs, resulting selectively appropriate for each one of the exotic BO tasks considered;
    \item presenting results on a set of global optimization test problems and for different dimensionalities of the search space.
\end{itemize}

\bmhead{Related works on collaborative/federated BO} Generally speaking, collaborative/federated black-box optimization is a task in which different \textit{agents} search for the global optimum of a black-box function through collaborative sequential queries. From a BO perspective, crucial challenges concern distributed computation, heterogeneity, and privacy (also known as \textit{data segregation}), because agents can just share their models or associated predictions but not observations. Many frameworks have been proposed, such as the three in \cite{al2024collaborative} or the one in \cite{yue2025collaborative}, but the idea of using the Wasserstein distance in this exotic BO task has just recently emerged \cite{zhan2025collaborative}.

\bmhead{Related works on batch BO} In batch BO, a set of solutions -- namely, a \textit{batch} -- is generated in parallel at every BO iteration. An important distinction is between \textit{synchronous} and \textit{asynchronous} approaches. In synchronous batch BO, the entire batch must be identified before starting the next BO iteration, whereas in asynchronous batch BO a new solution is searched for as soon as a single query is performed, in light of those still pending \cite{garnett2023bayesian}. In this paper, we focus only on synchronous batch BO, for which different strategies for building the batch have been proposed. According to \cite{garnett2023bayesian}, Thompson Sampling is the simplest strategy because it enables a trivial batch construction by sampling $q$ independent samples from the GP model, optimizing them separately as different surrogates of the true objective function and, finally, taking the resulting $q$ solutions as the new batch. Other strategies are based on sequential simulation -- basically applying a $q$-steps look-ahead schema -- or local penalization \cite{garnett2023bayesian}. The most important difference between collaborative/federated BO and batch BO approaches is that the last is usually based on a single GP model. However, we will show that batch BO can also be achieved by combining different GP models -- thanks to the WB barycenter of GPs -- without the need for sampling, look-ahead, or local penalization.

\bmhead{Related works on multi-fidelity BO}
Multi-fidelity BO (MFBO) methods optimize the black-box, expensive-to-evaluate, non-convex, multi-extremal objective function given the possibility to also query cheaper approximations of it, namely \textit{information sources}. A crucial assumption in MFBO is that the quality of approximation of each source, namely \textit{fidelity}, is known in advance and does not change over the search space. Consequently, the information sources can be hierarchically organized depending on their fidelity and the MFBO methods exploit this organization to outperform vanilla BO performed on the information source with the highest fidelity, that usually is the true function (aka \textit{ground-truth}). Depending on the source-specific observations, the information sources are modeled through as many independent GPs, then combined into a unique GP model, while the acquisition function suggests the next \textit{source-location} pair to query dealing with both the exploitation-exploration trade-off and the discrepancy between cheaper information sources and the \textit{ground-truth}. Pitfalls of MFBO methods have been reported in \cite{march2012provably,song2019general,mikkola2023multi}, arising whenever the basic assumptions are violated -- a quite frequent situation in real-life applications. Indeed, multiple information source Bayesian optimization (MISBO) has emerged to address MFBO's limitations, as widely discussed in \cite{candelieri2025multiple}. At the best of authors' knowledge, there are no MFBO or MISBO approaches using Wasserstein distance or WB.


\section{Background}\label{sec2}
\subsection{Wasserstein distance and Wasserstein Barycenter of probability measures}\label{sec2.1}

Here we summarize the background on the Wasserstein distance and the WB relatively to the scope of the paper; for a comprehensive discussion, the reader can refer to \cite{ollivier2014optimal,santambrogio2015optimal,peyre2019computational}.

Denote by $\mathcal{X}$ an arbitrary space and by $d:\mathcal{X}\times\mathcal{X}\rightarrow\mathbb{R}$ a \textit{ground metric} (i.e., a metric distance) on that space. Let $\mathbb{P}(\mathcal{X})$ be the set of Borel probability measures on $\mathcal{X}$, and $\mathcal{P}$ and $\mathcal{Q}$ two probability measures in $\mathbb{P}(\mathcal{X})$, then their $p$-Wasserstein distance is given by:
\begin{equation}
    \label{eq:1}
    W_p^p(\mathcal{P},\mathcal{Q})=\underset{\pi \in \Pi(\mathcal{P},\mathcal{Q})}{\inf}\; \int_{\mathcal{X}\times\mathcal{X}} d^p(x,x')\text{d}\pi(x,x')
\end{equation}
with $p\geq1$ and where $\Pi(\mathcal{P},\mathcal{Q})$ denotes the set of all probability measures in $\mathcal{X}\times\mathcal{X}$ having $\mathcal{P}$ and $\mathcal{Q}$ as marginals.\\

\bmhead{The 2-Wasserstein distance} If the ground-metric $d(\cdot,\cdot)$ is the Euclidean distance and $p=2$, we obtain the so-called 2-Wasserstein distance -- that is used in this paper -- and equation (\ref{eq:1}) can be rewritten as:
\begin{equation}
    \label{eq:2}
    W_2^2(\mathcal{P},\mathcal{Q})=\underset{\pi \in \Pi(\mathcal{P},\mathcal{Q})}{\min} \int_{\mathcal{X}\times\mathcal{X}} \|x-x'\|^2_2\text{d}\pi(x,x')
\end{equation}

which is intended as an Optimal Transport problem \cite{peyre2019computational}: the \textit{source} probability mass of $\mathcal{P}$ must be moved to match the \textit{target} probability mass of $\mathcal{Q}$ by entailing the minimum \textit{transportation cost}. Specifically, this cost is calculated according to the ground metric, while the constraint related to the match between the transported source and the target is denoted by $\pi \in \Pi(\mathcal{P},\mathcal{Q})$.\\

\bmhead{Wasserstein Barycenters of probability measures} Given a set of $M$ probability measures, their Wasserstein Barycenter $\bar{\mathcal{P}}$ is the solution to the following variational optimization problem:
\begin{equation}
    \label{eq:3}
    \bar{\mathcal{P}} \in \underset{\mathcal{P} \in \mathbb{P}(\mathcal{X})}{\arg \min} \sum_{m=1}^M \lambda_m W_2^2(\mathcal{P},\mathcal{P}_m)
\end{equation}
with $\lambda_m\geq0$ and $\sum_{m=1}^M\lambda_m=1$.\\

In simple terms, $\bar{\mathcal{P}}$ is the probability measure that minimizes the weighted sum of the Wasserstein distances from the $M$ given ones.
Clearly, different values of the weights $\lambda_1,...,\lambda_M$ lead to different WBs, given the same set of probability measures. That is why Wasserstein Barycenters (i.e., plural) or \textit{weighted} WB are the names adopted in the literature.\\

Problem (3) is convex\footnote{even if it is usually a difficult nonconvex problem in the case of a generic distance or discrete probability measures \cite{peyre2019computational}.} and a solution always exists; thus efficient solvers can be used. Moreover, the uniqueness of the solution of (3) was proved in \cite{agueh2011barycenters}. Finally, there exist some special cases for which the solution is simple or it has an explicit form, such as in the case of Gaussian Distributions (GDs), as discussed below.\\

\bmhead{The 2-Wasserstein distance between Gaussian Distributions} When the two GDs are considered, their 2-Wasserstein distance is simply given by
\begin{equation}
    \label{eq:4}
    W_2^2\left(\mathcal{N}(\mu_1,\Sigma_1),\mathcal{N}(\mu_2,\Sigma_2)\right) = \|\mu_1-\mu_2\|^2 + \mathcal{B}(\Sigma_1,\Sigma_2)^2
\end{equation}
where $\mathcal{B}$ is the Bures metric \cite{bures1969extension} between positive definite matrices \cite{forrester2016relating}, that is:
\begin{equation}
    \label{eq:5}
    \mathcal{B}(\Sigma_1,\Sigma_2) = \text{Tr}\left(\Sigma_1+\Sigma_2-2\left(\Sigma_1^\frac{1}{2} \Sigma_2, \Sigma_1^\frac{1}{2}\right)^\frac{1}{2}\right)
\end{equation}

In the case of centered GDs (i.e., $\mu_1 = \mu_2 = 0$), the 2-Wasserstein distance resembles the Bures metric. Moreover, if $\Sigma_1$ and $\Sigma_2 = $ are diagonal, the Bures metric is the Hellinger distance. Finally, in the commutative case, that is $\Sigma_1\Sigma_2=\Sigma_2\Sigma_1$, the Bures metric is equal to the Frobenius norm $\|\Sigma_1^{1/2}-\Sigma_2^{1/2}\|^2_{Frobenius}$.\\

When a set of $M$ independent GDs, is given, namely $\big\{\mathcal{N}(\mu_m,\Sigma_m)\big\}_{m=1:M}$, their weighted WB has an explicit form: it is a new GD, $\mathcal{N}(\mu,\Sigma)$, with
\begin{equation}
    \label{eq:6}
     \mu = \sum_{m=1}^M\lambda_m \mu_m;\;\;\Sigma=\sum_{m=1}^M \lambda_m \left(\Sigma^\frac{1}{2}\Sigma_m \Sigma^\frac{1}{2}\right)^\frac{1}{2}
\end{equation}

More important for the scope of this paper, in the case of $M$ univariate GDs, namely $\big\{\mathcal{N}(\mu_m,\sigma^2_m)\big\}_{m=1:M}$,the resulting WB is just a new univariate GD, $\mathcal{N}(\mu,\sigma^2)$, with
\begin{equation}
    \label{eq:7}
    \mu=\sum_{m=1}^M \lambda_m \mu_m;\;\;\sigma^2=\left(\sum_{m=1}^M\lambda_m\sigma_m\right)^2
\end{equation}
\\

From a statistical perspective, the normal random variable $A \sim \mathcal{N}(\mu,\sigma^2)$ is the sum of the $M$ independent normal random variables $\left\{A_m \sim \mathcal{N}\left(\lambda_m\mu_m,\left(\lambda_m\sigma_m\right)^2\right)\right\}_{m=1}^M$, that is $A=\sum_{m=1}^M A_m$.\\

\subsection{Wasserstein distance and the WB of GPs}
A GP regression model can be considered as an extension of a GD from scalar values to functions. Similarly to a GD, a GP is fully specified by its mean (function) $\mu(x)$ and covariance (function) $k(x,x')$, usually chosen among suitable kernel functions \cite{williams2006gaussian,gramacy2020surrogates}.

Every kernel function has its own hyperparameters $\theta \in \mathbb{R}^h$ and fitting a GP means tuning $\theta$ depending on a set of observations $D=\left\{(x^{(i)},y^{(i)})\right\}_{i=1:n}$, with $x^{(i)}\in\mathcal{X}$ and where $y^{(i)}$ is a possibly noisy observation of the black-box target function $f(x)$. It is usually assumed $y^{(i)}=f(x^{(i)})+\varepsilon^{(i)}$ where $\varepsilon\sim\mathcal{N}(0,\sigma_\varepsilon^2)$.

The GP fitting process is typically performed via Maximum Likelihood Estimation (MLE) or Maximum-A-Posteriori (MAP), which is basically a penalized MLE. The resulting GP's predictive (aka posterior) mean and variance are, respectively:
\begin{equation}
    \label{eq:8}
    \mu(x) = \mu_0(x) + \text{k}(x,\text{X})\left[\text{K}+\sigma_\varepsilon^2 \text{I}\right]^{-1} \left(\text{y}-\mu_o(\text{X})\right)
\end{equation}
\begin{equation}
    \label{eq:9}
    \sigma^2(x) = k(x,x) - \text{k}(x,\text{X})\left[\text{K}+\sigma_\varepsilon^2 \text{I}\right]^{-1} \text{k}(\text{X},x)
\end{equation}

\vspace{0.5cm}
where $\mu_0(x)$ is the GP's prior -- usually fixed to zero without loss of generality -- $\text{k}(x,\text{X})$ is a $n$-dimensional vector whose $i$th component is $k_\theta(x,x^{(i)})$, $\text{K}$ is the $n\times n$ kernel matrix whose entries are $\text{K}_{ij}=k_\theta(x^{(i)},x^{(j)})$, $\mu_0(\text{X})$ is an $n$-dimensional vector with the $i$th component equal to $\mu_0(x^{(i)})$, and finally $\text{k}(\text{X},x)=\text{k}(x,\text{X})^\top$.\\

The problem of computing the WB of GPs has been treated in \cite{mallasto2017learning} with a focus on learning from uncertain curves (i.e., GPs regression models) to explicitly combine the GPs into a unique model. This involves the computation of the following 2-Wasserstein distance between two GPs
\begin{equation}
    \label{eq:10}
    W^2_2(\mathcal{GP}_1,\mathcal{GP}_2) = d_2^2(\mu_1,\mu_2)+\text{Tr}\left(\text{K}_1+\text{K}_2-2\left(\text{K}_1^\frac{1}{2}\text{K}_2\text{K}_1^\frac{1}{2}\right)^\frac{1}{2}\right)
\end{equation}

The most important result in \cite{mallasto2017learning} is the explicit form of the WB of $M$ GPs, which is a new GP whose predictive mean $\bar{\mu}$ and kernel matrix $\bar{\text{K}}$ are:
\begin{equation}
    \label{eq:11}
     \bar\mu = \sum_{m=1}^M\lambda_m \mu_m;\;\;\bar{\text{K}}=\sum_{m=1}^M \lambda_m \left(\bar{\text{K}}^\frac{1}{2}\text{K}_m\bar{\text{K}}^\frac{1}{2}\right)^\frac{1}{2}
\end{equation}
with weights $\lambda_1,...,\lambda_M$ given.

Just like for GDs, the existence and uniqueness of the WB of GPs are both guaranteed, under the assumption that the WB is nondegenerate. As remarked in \cite{mallasto2017learning}, it is still a conjecture that the barycenter of non-degenerate GPs is nondegenerate \cite{masarotto2019procrustes}, but this holds in the finite-dimensional case of GDs.
Although we have a strict analogy to the WB of GDs, computing the WB of GPs is not trivial due to the calculation of $\bar{\text{K}}$. In fact, it requires a fixed-point iteration scheme -- for which there is, anyway, a convergence proof \cite{mallasto2017learning} -- leading to approximate the WB of $M$ GPs through the WB of a set of GDs converging to those GPs.

In the following, we explain how these limitations can be overcome, in BO, by fully exploiting, at the acquisition function level, the fact that GP's predictions at every $x$ are always univariate GDs, independently on the dimensionality of the search space.


\section{Weighted WB of GPs (W2BGP) for exotic BO tasks}
\subsection{Acquisitions functions for Wasserstein Barycenter of GPs}
Along with the GP regression model, the \textit{acquisition function} is the other key component of BO, driving the choice of the next solution to evaluate depending on the current GP \cite{archetti2019bayesian,garnett2023bayesian}. Generally speaking, the acquisition function quantifies the \textit{utility} in selecting a solution $x \in \mathcal{X}$ while balancing between exploitation (local search) and exploration (global search). Formally, the next solution to be evaluated is obtained as:
\begin{equation}
    \label{eq:12}
    x^{(n+1)} \in \underset{x \in \mathcal{X}}{\arg \max}\; \alpha\left(x;\mu(x),\sigma(x)\right)
\end{equation}
where $n$ is the number of current observations stored in $D=\left\{(x^{(i)},y^{(i)})\right\}_{i=1:n}$.\\

Most of the acquisition functions consist of some combination of $\mu(x)$ and $\sigma(x)$ and are evaluated point-wise along gradient-based procedures or derivative-free optimization methods. For every candidate solution $x$ the underlying assumption is that $f(x)\sim\mathcal{N}\left(\mu(x),\sigma^2(x)\right)$, that is a random variable from a univariate GD. According to what was discussed in Section \ref{sec2.1}, and following from (\ref{eq:7}), this means that the (weighted) WB of the $M$ independent univariate GDs at $x$ -- associated with $M$ GPs -- is a new univariate GD with mean and variance equal to:
\begin{equation}
    \label{eq:13}
    \bar\mu(x)= \sum_{m=1}^M \lambda_m \mu_m(x);\;\;\bar\sigma^2(x)=\left(\sum_{m=1}^M\lambda_m\sigma_m(x)\right)^2
\end{equation}

This leads to a relevant simplification: we no longer need to compute $\bar{\text{K}}$, as in \cite{mallasto2017learning}, but just computing the WB between $M$ univariate GDs (i.e., a really cheap calculation) at every $x$ that is needed (e.g., along iterations of a gradient-based method or for pools of solutions in evolutionary optimization methods).\\

\begin{figure}[h!]
    \centering
    \includegraphics[width=0.9\linewidth]{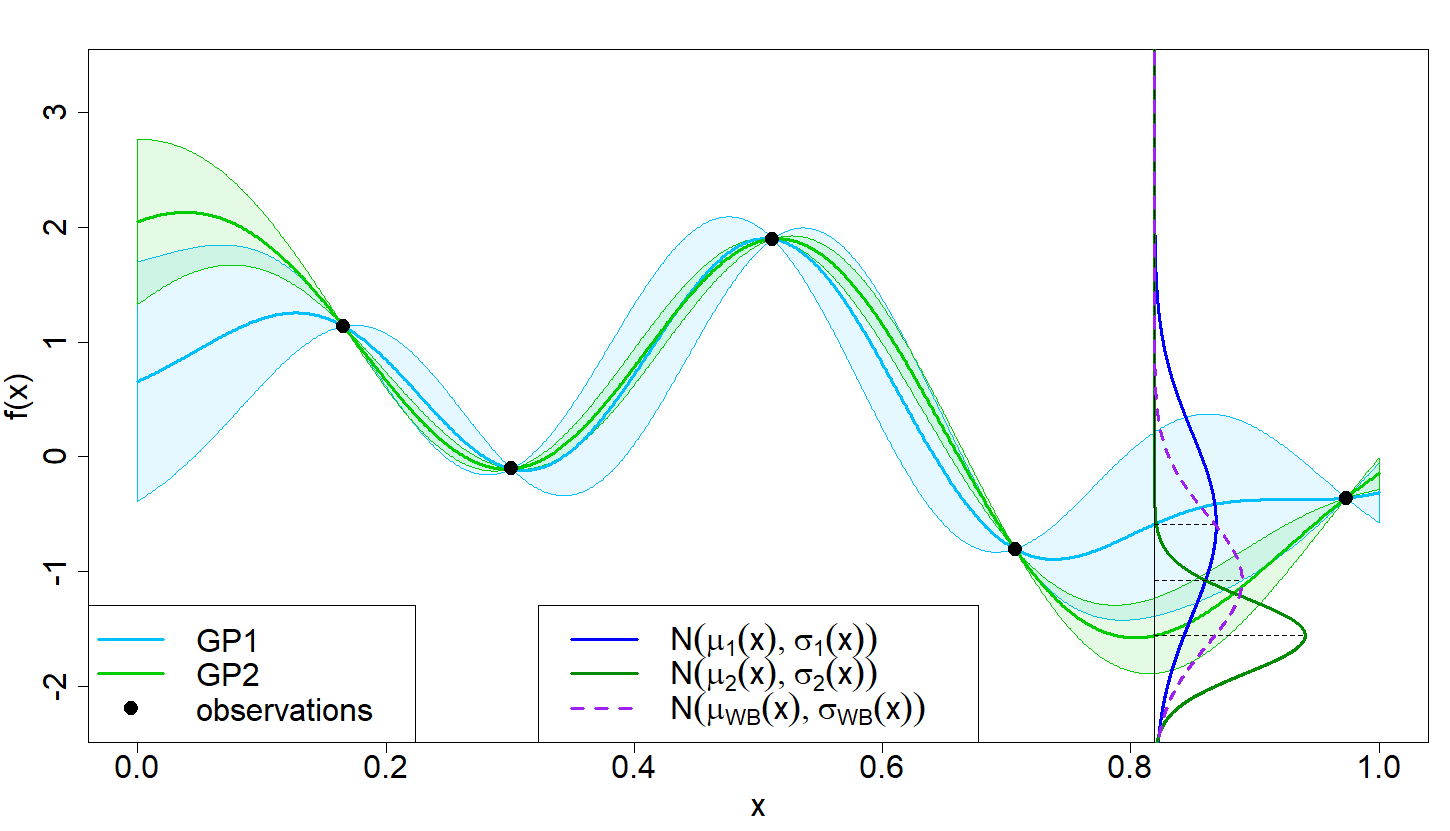}
    \caption{Two GPs fitted on the same set of observations. At $x=0.81$ is it possible to observe: the normal distribution associated to the first GP, namely $\mathcal{N}(\mu_1(0.81),\sigma_1(0.81))$ (blue solid curve), normal distribution associated to the second GP, namely $\mathcal{N}(\mu_2(0.81),\sigma_2(0.81))$ (green solid curve), and the resulting WB, with $\lambda_1=\lambda_2=0.5$ (dashed purple curve).}
    \label{fig:1}
\end{figure}

To make this crucial concept very clear, we illustrate an easy 1-dimensional example: Figure \ref{fig:1} shows two different GPs -- using different kernels or the same kernel but with different values of the hyperparameters -- fitted on the same set of observations. At $x=0.81$ the two univariate normal distributions given by the two GPs, namely $\mathcal{N}(\mu_1(0.81),\sigma_1(0.81))$ and $\mathcal{N}(\mu_2(0.81),\sigma_2(0.81))$, are depicted (they have been preliminary rescaled for a pretty visualization). The resulting WB, with $\lambda_1=\lambda_2=0.5$, is also reported (i.e., dashed purple curve).

\begin{figure}[h!]
    \centering
    \includegraphics[width=0.9\linewidth]{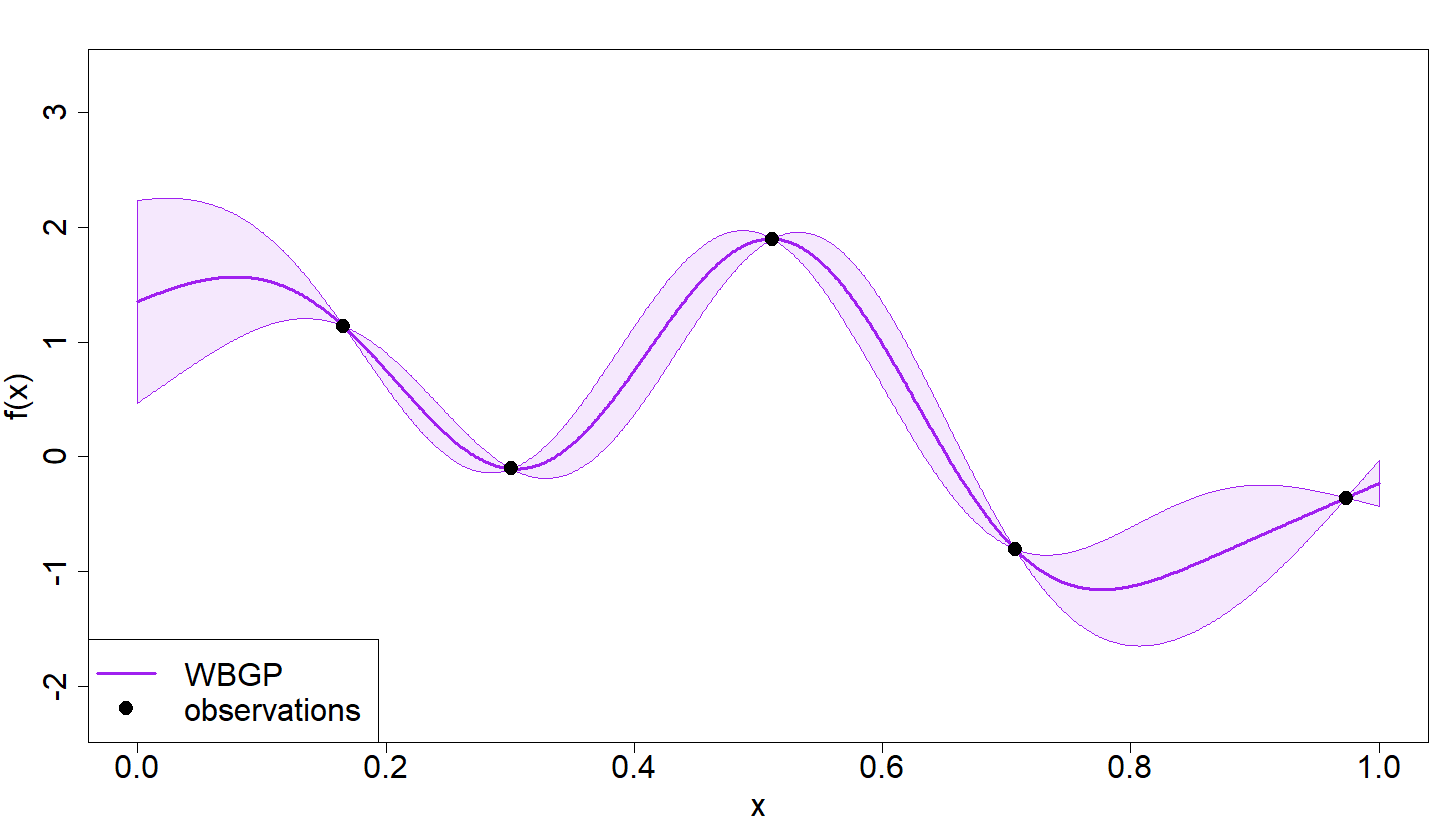}
    \caption{The resulting WB of the two GPs in Figure \ref{fig:1}, with $\lambda_1=\lambda_2=0.5$.}
    \label{fig:2}
\end{figure}

Computing the WB at all possible locations $x\in\mathcal{X}$ leads to the WB of the GPs depicted in Figure \ref{fig:2}.
 
\newpage 
As follows, we present how three of the most widely known and used acquisition functions change when the weighted WB of $M$ GPs -- namely W2BGP -- is adopted as a probabilistic surrogate model. As a reference problem, we consider the global minimization of a black-box, expensive-to-evaluate, non-convex function $f(x)$:
\begin{equation}
    \label{eq:14}
    \underset{x\in\mathcal{X}}{\min}\; f(x)
\end{equation}
\newline

\noindent
\textbf{W2BGP Lower Confidence Bound.} \textit{The Lower Confidence Bound of the weighted WB of $M$ different GPs is the convex combination of the their individual Lower Confidence Bound functions.}
\begin{equation}
    \bar\alpha^{LCB}(x;\lambda) = \sum_{m=1}^M\lambda_m \alpha^{LCB}_m(x)
\end{equation}
\newline

\noindent
The proof is quite trivial, directly stemming from (\ref{eq:13}):
\begin{equation}
    \begin{split}
        \bar\alpha^{LCB}(x;\lambda) & = \bar\mu(x;\lambda) - \beta \bar\sigma(x;\lambda) = \sum_{m=1}^M \lambda_m \mu_m(x) - \beta \sum_{m=1}^M \lambda_m \sigma_m(x)=\\
        & = \sum_{m=1}^M \lambda_m \big(\mu_m(x) - \beta \sigma_m(x)\big) = \sum_{m=1}^M \lambda_m \alpha_m^{LCB}(x).
    \end{split}
\end{equation}
\newline

\noindent
\textbf{W2BGP Probability of Improvement.} \textit{Probability of Improvement of the weighted WB of $M$ different GPs is the CDF (cumulative density function) of a sum of independent univariate normal random variables.}
\begin{equation}
    \begin{split}
        \bar \alpha^{PI}(x;\lambda) & = \Phi\left(\frac{y^+-\bar\mu(x;\lambda)}{\bar\sigma(x;\lambda)}\right) = \Phi\left(\frac{\sum_{m=1}^M (y^+-\lambda_m\mu_m(x))}{\sum_{m=1}^M \lambda_m \sigma_m(x)}\right)
    \end{split}
\end{equation}
where $y^+=\underset{i=1,...,n}{\min}\left\{y^{(i)}\right\}$ is the best value observed so far, also known as \textit{best-seen}.
\newline

\noindent
\textbf{W2BGP Expected Improvement.} \textit{The Expected Improvement of the weighted WB of $M$ different GPs entails the computation of the CDF and PDF (probability density function) of $M$ independent univariate normal random variables. Specifically, the WB's PDF is the convolutions of the single PDFs}.
\begin{equation}
    \begin{split}
    \bar \alpha^{EI}(x;\lambda) & = \left(y^+ - \bar\mu(x;\lambda)\right)\Phi\left(\frac{y^+ - \bar\mu(x;\lambda)}{\bar\sigma(x;\lambda)}\right) + \bar\sigma(x;\lambda) \phi\left(\frac{y^+ - \bar\mu(x;\lambda)}{\bar\sigma(x;\lambda)}\right) =\\
    & = \left(\sum_{m=1}^M (y^+ - \lambda_m\mu_m(x)) \right) \Phi\left(\frac{\sum_{m=1}^M (y^+ - \lambda_m \mu_m(x)) }{\sum_{m=1}^M \bar\sigma_m(x)}\right) + \\
    & + \left(\sum_{m=1}^M (y^+ - \lambda_m\sigma_m(x)) \right)\phi\left(\frac{\sum_{m=1}^M \lambda_m\mu_m(x)}{\sum_{m=1}^M\lambda_m\sigma_m(x)}\right).
    \end{split}
\end{equation}
if $\bar\sigma(x;\lambda)>0$ (i.e., $\sigma_m(x)>0\;\forall\;m\in\{1,...,M\}$), otherwise $\bar\alpha^{EI}(x;\lambda)=0$.\\

\subsection{W2BGP for Collaborative/Federated BO}
In collaborative/federated BO, $M$ agents optimize a target function by collaborating while guaranteeing data segregation. Specifically, every agent cannot share its own observations with the others, and collaboration is achieved by sharing models or, even better, just their predictions.
Formally, every agent has its own set of observations, 
$D_m=\left\{\left(x_m^{(i)},y_m^{(i)}\right)\right\}_{i=1:n_m}$, with $m=1,...,M$. Without loss of generality, it is convenient to consider $n_m=n, \forall\; m$. Depending on its own $D_m$, every agent fits its associated GP model, whose predictive mean and uncertainty are denoted by $\mu_m(x)$ and $\sigma_m(x)$.

Collaboration is usually managed by a central system that provides each of the $M$ agents with the next solution to evaluate, by privately combining the predictions of the individual agents. In this paper, the central system generates the next solutions via W2BGP's LCB by using a different weight vector for each agent. Formally,
\begin{equation}
    x_m^{(n+1)} \in \underset{x \in \mathcal{X}}{\arg \min}\; \bar{\alpha}^{LCB}\left(x;\lambda^{[m]}\right), \; \forall\;m\in\{1,...,M\} 
\end{equation}

After all $\left\{x^{(n+1)}_m\right\}_{m=1:M}$ have been individually and separately evaluated, each agent privately updates its own set of observations $D_m \leftarrow D_m \cup \left\{\left(x_m^{(n+1)},y_m^{(n+1)}\right)\right\}$, and then $n \leftarrow n+1$. As usual, the process is iterated until some termination criterion is met.\\

\noindent
In our experiments, we are going to consider the following three weighting schemes for the W2BGP:
\begin{itemize}
    \item \textbf{self-confident weighting schema.} For every agent, the central system creates a W2BGP that relies on that agent's GP model for a half and on the other agent's GP models for the remaining half, uniformly. In formal terms:
    \begin{equation}
        \lambda^{[m]}: \lambda^{[m]}_i =
            \begin{cases}
            0.5\quad\quad\quad\quad \text{ if } i=m\\
            0.5/(M-1) \text{ otherwise } 
            \end{cases} 
    \end{equation}

    \item \textbf{equal-weights schema.} The central system creates a unique W2BGP by giving the same weight to every agent's GP model. Formally:
    \begin{equation}
        \lambda^{[m]}: \lambda^{[m]}_i=1/M\;\forall\; m,i\in\{1,...,M\}
    \end{equation}

    Important, this weighting schema leads the central system to provide the same query to all agents, unbeknownst to them. Therefore, the $M$ GPs will become more and more similar along the sequential series.\\ 

    \item \textbf{uncooperative weighting schema.} The central system is not needed in this case: every agent works without collaborating with the others. This means that not only observations but also GP models are not shared, and the task resembles running multiple \textit{vanilla} BO experiments separately and independently. In terms of WB's weights this means
    \begin{equation}
        \lambda^{[m]}: \lambda^{[m]}_i =
            \begin{cases}
            1 \text{ if } i=m\\
            0 \text{ otherwise } 
            \end{cases} 
    \end{equation}\\
\end{itemize}

Figure \ref{fig:1} offers a graphical representation of the functioning of the three weighting schemes for the W2BGP in collaborative/federated BO.

\begin{figure}[H]
    \centering
    \includegraphics[width=0.48\linewidth]{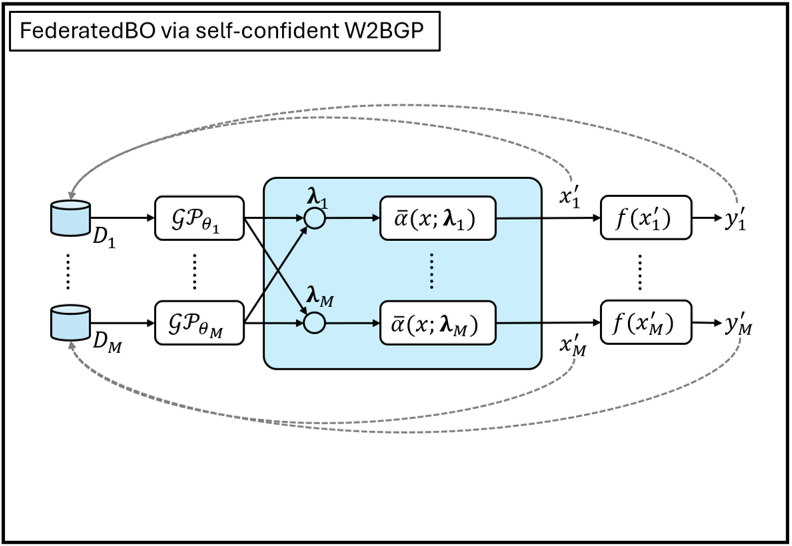}
    \includegraphics[width=0.48\linewidth]{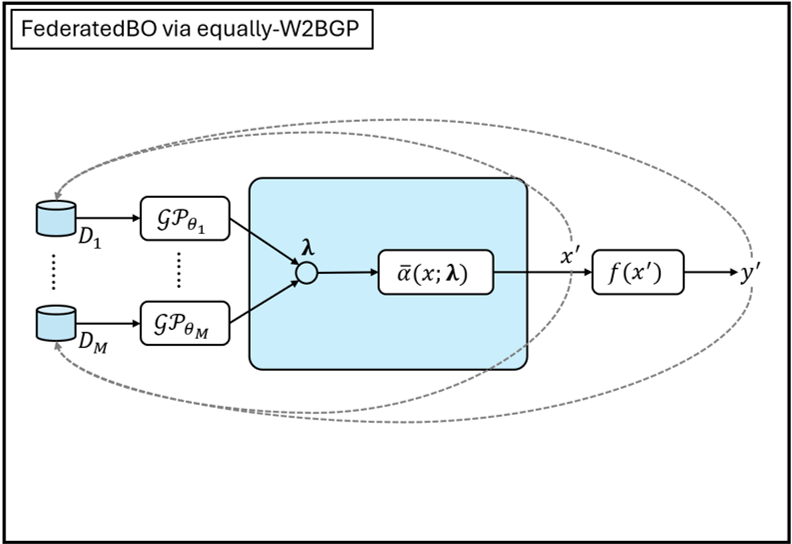}
    \includegraphics[width=0.48\linewidth]{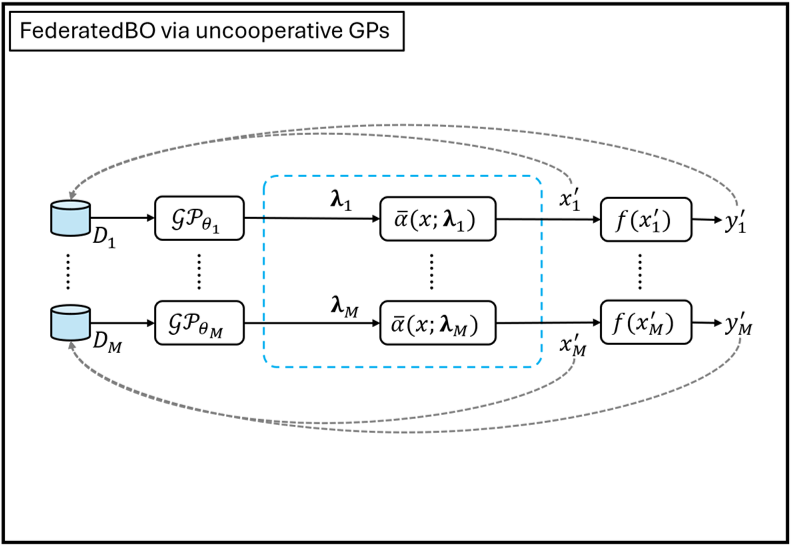}   
    \caption{The three weighting scheme for the W2BGP in collaborative/federated BO: self-confident weighting schema (up-left), equally-weighted schema (up-right), and uncooperative weighting schema (bottom).}
    \label{fig:3}
\end{figure}

\subsection{W2BGP for batch BO}
Contrary to the collaborative/federated BO, in batch BO there is no privacy issue. Thus, we have only one dataset of observations, that is $D=\{(x^{(i)},y^{(i)})\}_{i=1:n}$. In our proposal, we use a pool of $M$ different GPs to -- synchronously -- create our batch at every BO iteration, where the GPs are different because they rely on as many different kernels. Every GP model will provide its own predictive mean and uncertainty, namely $\mu_m(x;D,k_m)$ and $\sigma_m(x;D,k_m)$, with $k_m$ referring to the kernel type.

In batch BO, the two most interesting weighting schemes are \textit{self-confident} and \textit{uncooperative}, because they provide more than one query per iteration. A relevant and interesting difference between these two schemes and other state-of-the-art batch BO algorithms is with respect to the size $q$ of the batch. In other methods $q$ is fixed in advance and kept constant throughout iterations, while in our case the size of the batch \textit{should} be $q=M$ but, if some GP becomes close to another -- despite their different kernels -- then we will come up with $q<M$. More important, $q\rightarrow 1$ denotes convergence between GP models, which means that all become similar in terms of predictive mean and uncertainty. This behavior can emerge in our batch BO method because the different GP models share the same set of observations, contrary to the previous collaborative/federated BO task.

\begin{figure}[h!]
    \centering
    \includegraphics[width=0.49\linewidth]{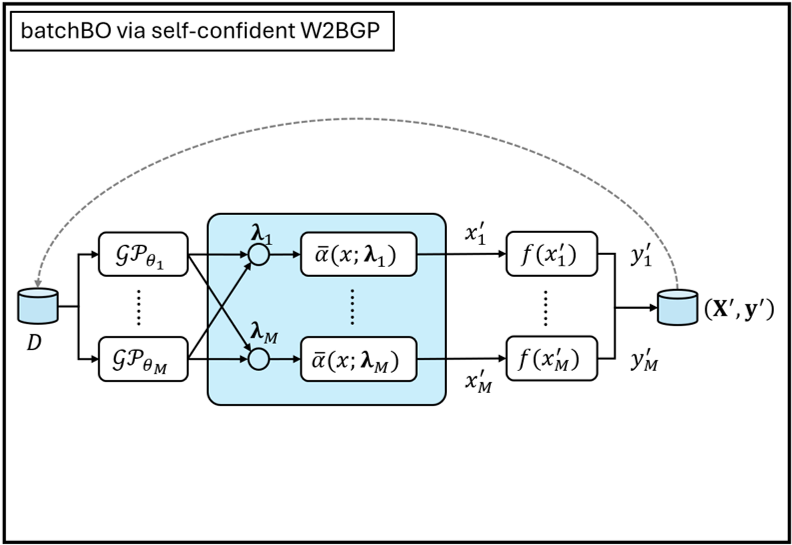}
    \includegraphics[width=0.49\linewidth]{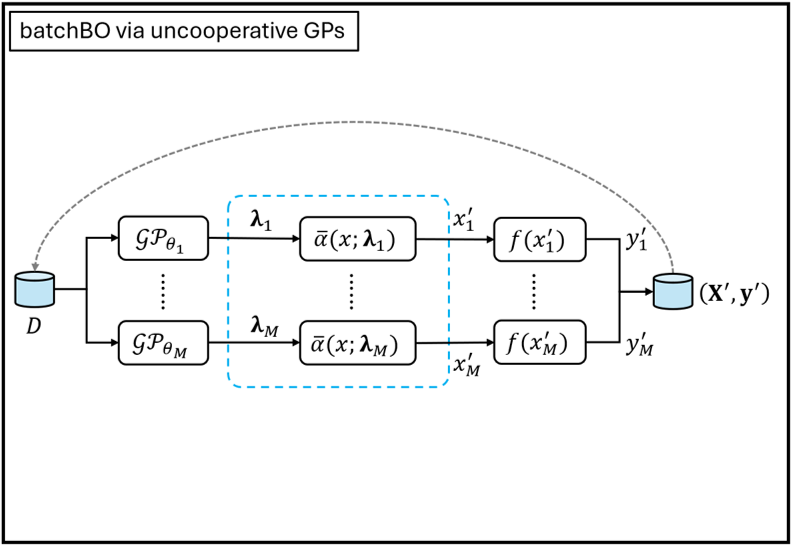}
    \caption{Graphical representation of the self-confident (left) and uncooperative (right) weighting schemes for batch BO. Equally-weighted schema is just model averaging and it is omitted in the figure.}
    \label{fig:4}
\end{figure}

Finally, it is important to clarify that using the equally weighted scheme for WB is no longer batch BO, but it resembles BO with simple model averaging \cite{garnett2023bayesian}. Thus, it is omitted in Figure \ref{fig:4}, which just offers a graphical representation of the functioning of the self-confident and uncooperative weighting schemes for WB in batch BO.

\subsection{W2BGP for multi-fidelity BO}
As for batch BO, and contrary to collaborative/federated BO, in multi-fidelity BO there is no data privacy issue. Moreover, this is the setting in which a reasonable weighting schema for the W2BGP would be easy to determine: it is sufficient to set the value of each weight proportionally to the fidelity of the associated information source.
Similarly to the previous two tasks, observations are collected in $M$ disjoint datasets, where each dataset is denoted with $D_m=\left\{(x_m^{(i)},y_m^{(i)})\right\}_{i=1:n_m}$, but,
in this case, we cannot assume $n_m=n, \forall\;m=1,...,M$ because the number of queries on each information source depends on the sequential optimization process. More precisely, at each iteration of every MFBO method, a \textit{location-source} pair is selected to be queried. This means that the sets $\{D_m\}_{(m=1:M}$ are not updated in parallel as in collaborative/federated BO and (our) batch BO.

In MFBO the $M$ GPs, separately modeling the different information sources, are combined (aka fused) into a unique GP model: usually the GPs share the same kernel, but anyway offer different predictive mean and uncertainty depending on their specific observation datasets, leading to $\mu_m(x)$ and $\sigma_m(x)$. In this paper, the proposed to use the W2BGP as a principled combination model.

For the acquisition function, we took inspiration from our previous papers \cite{candelieri2025multiple,candelieri2024fair} and propose here the following variants to specifically account for the W2BGP:
\begin{equation}
    m',x' \in \underset{\substack{x \in \mathcal{X}\\m\in\{1,...,M\}}}{\arg \min}\; \frac{ y^+ -\Big[\bar\mu(x;\lambda) - \beta \bar\sigma(x;\lambda)\Big]}{c_ms\Big[\big(\mu_m(x)-\bar\mu(x;\lambda)\big)^2 - \big(\sigma_m(x)-\bar\sigma(x;\lambda)\big)^2 \Big]}
    \label{eq:mfbo_acq}
\end{equation}
where the numerator is the most optimistic improvement with respect to the best value observed so far on the ground-truth, namely $y^+$, while at the denominator $c_m$ is the cost of querying the $m$th information source, and the other term is the Wasserstein distance between $\mathcal{N}(\mu_m(x),\sigma_m^2(x))$ and $\mathcal{N}(\bar\mu(x;\lambda),\sigma^2(x;\lambda))$, quantifying the discrepancy, at $x$, between the GP model of the $m$th information source and the W2BGP, with $\lambda$ as its weight vector. Then, $D_{m'}\leftarrow D_m \cup \{(x',f_{m'}(x')\}$ and $n\leftarrow n+1$.

Similarly to the other two exotic BO tasks, three weighting schemes have been considered and compared, but slightly different from the previous ones. The difference is motivated by the specific task that, as already mentioned, is well suited to define reasonable weighting schemes for the W2BGP. Specifically, we considered:
\begin{itemize}
    \item \textbf{fidelities as weights}: this is the most trivial weighting schema for the WB of GPS, with every $\lambda_m$ equal to the fidelity of the associated $m$th information source.
    \item \textbf{rescaled weights}: in this weighting schema the weights are still proportional to the fidelity of the associated information sources, but rescaled according to the rule: $\lambda_m = 0.75 (0.25)^{m-1}$, with $m=1,...M$.
    \item \textbf{equal weights}: as usual, the basic weighting schema which ignores the relevance of the different information sources; in fact, $\lambda_m=1/M$ for every $m=1,...M$.    
\end{itemize}


\section{Experiments and results}\label{sec:4}
\subsection{Experimental setting}
In our experiments, we want to empirically prove that W2BGP enables a unifying framework for solving those exotic BO tasks that require the combination of different GPs as a core operation.
According to this statement, we are not here interested in comparing our approach against specific BO algorithms that selectively solve just one of the three tasks. Our aim is to analyze whether and how the weights of the W2BGP can be reasonably tuned, depending on the task, to obtain good performances.\\

\bmhead{Performance metric} For all the three considered tasks, sample efficiency has been measured according to the Area Under the Gap metric Curve (AUGC), a measure already adopted in \cite{candelieri2024mle}, with the Gap metric \cite{huang2006global} defined as follows, in the case of a minimization problem:
\begin{equation}
    G_n = \frac{y_0-y^+}{y_0-y^*}
\end{equation}
where $y^+$ is usually called \textit{best seen}, that is the best value observed so far. Thus, in our global minimization setting $y^+=\min_{i=1:n}\left\{y^{(i)}\right\}$.
The gap metric represents how much the optimization process improves with respect to the best initial solution and according to the optimal value, and it is well-suited for comparing different algorithms -- even across different problems -- given the knowledge of the optimum, thus, it is usually adopted in the case of test problems.\
The AUGC is then computed as:
\begin{equation}
    AUGC = \frac{1}{N} \sum_{n=1}^N G_n = \frac{1}{N} \sum_{n=1}^N \frac{y_0-y^*}{y_n-y^*}
\end{equation}
and it is analogous to the famous AUC -- Area Under the (Receiving Operating) Curve -- in ML. The AUGC quantifies, in a unique scalar value, how much and how fast an optimization algorithm is able to improve from the initial solution and to get close to the optimum.

\bmhead{Test problems} For collaborative/federated BO and batch BO, we have considered eight test problems with $d=1$, six test problems with $d=2$, and eight test problems with $d$ varying from 3 up to 20. For multi-fidelity BO, we have considered a set of eight test problems, specifically designed as multi-fidelity benchmarks and proposed in \cite{mainini2022analytical}.  In this case, the dimensionality $d$ of the search space ranges from $1$ to $5$, as suggested in the quoted paper. Details of all test problems considered in the paper are thoroughly documented in the Appendix.

\bmhead{BO settings} For collaborative/federated BO and batch BO, we have set $M=4$, that is the number of different kernel functions used to (a) model agents in collaborative/federated BO and (b) identify solutions for each batch in batch BO. Specifically, the following kernel functions have been considered: Exponential (aka Laplacian), Squared Exponential (aka Gaussian), Matérn 3/2, and Matérn 5/2.\\
For MFBO, $M$ is equal to the number of fidelity levels (i.e., the number of information sources) in each multi-fidelity test problem. The Squared Exponential kernel has been kept fixed for all the GPs and the test problems.\\
For all three exotic BO tasks, the initial set of random solutions has been defined via Latin Hypercube Sampling (LHS); the size of this initial set is given by $n_0=\max\{d+1,\min\{2d,10\}\}$. The total number of queries (including the initial ones) is given by $N=\min\{30d,150\}$.\\
To mitigate the impact of random initialization, we performed 30 independent runs and, to achieve a fair comparison, the initial set of random solutions is shared among the different weighting schemes of the W2BGP model, for each independent run and for each exotic BO task.\\

The results for the three exotic tasks considered in this paper are reported and discussed separately in the following three subsections.\\

\subsection{W2BGP for Collaborative/Federated BO: results}
In this section, we summarize the most relevant results on the application of W2BGP in collaborative/federated BO. Specifically, we report the AUGC in Table \ref{tab:1} and the best seen at the end of the optimization process in Table \ref{tab:2}, respectively. Values are median and standard deviation on 30 independent runs.

\bmhead{Result \#1} \textit{The equal weighting schema (that is, simple model averaging) is the worst choice for W2BGP to target a collaborative/federated BO task.}

Although the equal weighting schema (i.e., model averaging) could represent the most intuitive and a quite reasonable choice for many ML tasks, especially whenever prior knowledge is not available, applying it to W2BGP for collaborative/federated BO leads to significantly poor performance, in terms of both AUGC and best seen.

\bmhead{Result \#2} \textit{The self-confident weighting schema is the most efficient choice in collaborative/federated BO.}

As reported in Table \ref{tab:1}, the AUGC for the self-weighting schema is the largest for almost all the test problems considered, resulting in a higher sample efficiency compared to the other two weighting schemes. On the other hand, the uncooperative weighting schema was competitive in terms of best seen (Table \ref{tab:2}). In other terms, if every agent includes some knowledge from the others to make its own decision, this leads the overall collaborative/federated system to quickly converge to an optimal solution, even if it could be slightly worse -- in some cases -- than a set of completely independent, uncooperative BO agents. A possibility -- not investigated in this paper -- might be to switch from self-confident to uncooperative schema when the number of queries is approaching the maximum allowed number.\\

Just as a relevant example, in Figure \ref{fig:2} we report the Gap metric curves of the three weighting schemes for the \textit{alpine01} test problem, with $d=\{5,10,20\}$. The entire set of charts for all the test problems is reported, for completeness, in the Appendix. The better performance provided by the self-confident weighting schema is clear, especially with $d$ increasing.

\begin{table}[h!]
    \centering
    \resizebox{\columnwidth}{!}{%
    \begin{tabular}{l|l|ccc|cc}
        \hline\hline        
        & \textbf{Test} & \textbf{self-confident} & \textbf{equally} & \textbf{uncooperative} & \textbf{U test} & \textbf{U test} \\
        d & \textbf{problem} & \textbf{W2BGPBO} & \textbf{W2BGPBO} & \textbf{parallel GPBO} & $p$\textbf{-value} & $p$\textbf{-value} \\
        & & \textcircled{a} & \textcircled{b} & \textcircled{c} & \textcircled{a} vs \textcircled{b} & \textcircled{a} vs \textcircled{c} \\
        
        \hline\hline
        1 & problem\_02 & \textbf{0.8752} (0.0594) & 0.8268 (0.0685) & 0.8611 (0.0716) & \textit{0.000} & 0.824 \\
        1 & problem\_03 & 0.6832 (0.2155) & 0.4110 (0.3054) & \textbf{0.7070} (0.2483) & \textit{0.001} & 0.984 \\
        1 & problem\_05 & \textbf{0.8501} (0.1404) & 0.6570 (0.2122) & 0.8196 (0.1395) & \textit{0.000} & 0.641 \\
        1 & problem\_07 & \textbf{0.8778} (0.0676) & 0.7914 (0.1146) & 0.8591 (0.0577) & \textit{0.000} & 0.318 \\
        1 & problem\_11 & \textbf{0.8693} (0.2561) & 0.7567 (0.2211) & 0.8630 (0.1145) & \textit{0.001} & 0.341 \\
        1 & problem\_14 & \textbf{0.8258} (0.1349) & 0.4793 (0.2403) & 0.7867 (0.2054) & \textit{0.000} & 0.198 \\
        1 & problem\_15 & 0.8548 (0.2064) & 0.7418 (0.2392) & \textbf{0.8911} (0.1012) & \textit{0.001} & \textit{0.025} \\
        1 & problem\_22 & 0.8741 (0.0725) & 0.6766 (0.1667) & \textbf{0.8780} (0.1494) & \textit{0.000} & 0.746 \\
        \hline
        2 & alpine01 & \textbf{0.8629} (0.0768) & 0.7631 (0.1766) & 0.8017 (0.1475) & \textit{0.000} & \textit{0.002} \\
        2 & bird & \textbf{0.8313} (0.0740) & 0.7540 (0.1961) & 0.7794 (0.0793) & \textit{0.000} & \textit{0.043} \\
        2 & michalewicz & \textbf{0.8643} (0.0723) & 0.7720 (0.1830) & 0.8162 (0.1017) & \textit{0.000} & \textit{0.000} \\
        2 & styblinskiTang & \textbf{0.8309} (0.0819) & 0.7233 (0.1064) & 0.8243 (0.0811) & \textit{0.000} & 0.688 \\
        2 & ursem03 & \textbf{0.7635} (0.1676) & 0.4381 (0.2224) & 0.5607 (0.2571) & \textit{0.000} & \textit{0.000} \\
        2 & ursemWaves & 0.9376 (0.0623) & 0.8545 (0.1305) & \textbf{0.9476} (0.0580) & \textit{0.003} & 0.761 \\
        \hline
        3 & hartmann3 & \textbf{0.9271} (0.0268) & 0.9047 (0.1160) & 0.9220 (0.0231) & \textit{0.001} & 0.472 \\
        6 & hartmann6 & 0.6306 (0.0803) & 0.5772 (0.0802) & \textbf{0.6551} (0.0684) & \textit{0.000} & 0.360 \\
        \hline
        5 & alpine01 & \textbf{0.9030} (0.0662) & 0.8404 (0.1225) & 0.8418 (0.0889) & \textit{0.000} & \textit{0.000} \\
        10 & alpine01 & \textbf{0.8575} (0.0324) & 0.6757 (0.0774) & 0.7757 (0.0672) & \textit{0.000} & \textit{0.000} \\
        20 & alpine01 & \textbf{0.7564} (0.0468) & 0.5717 (0.0876) & 0.6226 (0.0936) & \textit{0.000} & \textit{0.000} \\
        \hline
        5 & styblinskiTang & \textbf{0.7949} (0.0701) & 0.7042 (0.0986) & 0.7541(0.0705) & \textit{0.000} & \textit{0.008} \\
        10 & styblinskiTang & \textbf{0.6259} (0.1411) & 0.4863 (0.1349) & 0.5647 (0.1054) & \textit{0.001} & 0.119 \\
        20 & styblinskiTang & \textbf{0.5699} (0.1143) & 0.4260 (0.1158) & 0.3449 (0.0830) & \textit{0.000} & \textit{0.000} \\
        \hline\hline
    \end{tabular}%
    }
    \caption{AUGC on collaborative/federated BO: median (standard deviation) on 30 independent runs: the higher the better (in bold, the highest median value for each test problem).}
    \label{tab:1}
\end{table}

\clearpage

\begin{table}[h!]
    \centering
    \resizebox{\columnwidth}{!}{%
    \begin{tabular}{l|l|ccc|cc}
        \hline\hline        
        & \textbf{Test} & \textbf{self-confident} & \textbf{equally} & \textbf{uncooperative} & \textbf{U test} & \textbf{U test} \\
        d & \textbf{problem} & \textbf{W2BGPBO} & \textbf{W2BGPBO} & \textbf{parallel GPBO} & $p$\textbf{-value} & $p$\textbf{-value} \\
        & & \textcircled{a} & \textcircled{b} & \textcircled{c} & \textcircled{a} vs \textcircled{b} & \textcircled{a} vs \textcircled{c} \\
        
        \hline\hline
        1 & problem\_02 & \textbf{-1.8996 (0.0000)} & \textbf{-1.8996 (0.0000)} & \textbf{-1.8996 (0.0000)} & 0.233 & 1.000 \\
        1 & problem\_03 & \textbf{-12.0293} (0.6035) & -11.9874 (1.7414) & -12.0064 (0.3994) & \textit{0.004} & \textit{0.034} \\
        1 & problem\_05 & \textbf{-1.4891 (0.0001)} & -1.4890 (0.0470) & \textbf{-1.4891} (0.0115) & \textit{0.003} & \textit{0.018} \\
        1 & problem\_07 & \textbf{-1.6013 (0.0000)} & \textbf{-1.6013} (0.0062) & \textbf{-1.6013 (0.0000)} & 0.124 & 0.346 \\
        1 & problem\_11 & \textbf{-1.5000 (0.0000)} & \textbf{-1.5000} (0.0001) & \textbf{-1.5000 (0.0000)} & 0.061 & \textit{0.048} \\
        1 & problem\_14 & \textbf{-0.7887 (0.0001)} & -0.7886 (0.0909) & \textbf{-0.7887} (0.0005) & \textit{0.002} & 1.000 \\
        1 & problem\_15 & \textbf{-0.0355} (0.0048) & \textbf{-0.0355} (0.0015) & \textbf{-0.0355 (0.0000) } & 0.665 & \textit{0.021} \\
        1 & problem\_22 & \textbf{-1.0000 (0.0000)} & -0.9999 (0.0146) & \textbf{-1.0000} (0.0001) & \textit{0.002} & 0.930 \\
        \hline
        2 & alpine01 & \textbf{0.0016} (0.0078) & 0.0033 (0.0224) & 0.0028 (0.0069) & \textit{0.026} & \textit{0.125} \\
        2 & bird & -106.7633 (0.0023) & -106.7560 (7.9267) & \textbf{-106.7644} (0.0202) & \textit{0.000} & 0.358 \\
        2 & michalewicz & \textbf{-1.8013 (0.0001)} & -1.8012 (0.1079) & \textbf{-1.8013} (0.0005) & \textit{0.003} & 0.133 \\
        2 & styblinskiTang & -78.3236 (0.0864) & -78.3132 (0.1032) & \textbf{-78.3323} (0.0001) & 0.328 & \textit{0.000} \\
        2 & ursem03 & \textbf{-2.9954} (0.1665) & -2.9688 (0.4005) & -2.9810 (0.2463) & \textit{0.000} & \textit{0.000} \\
        2 & ursemWaves & -7.3067 (0.0009) & -7.3043 (0.0011) & \textbf{-7.3070} (0.0009) & \textit{0.001} & 0.578 \\
        \hline
        3 & hartmann3 & -3.8627 (0.0006) & -3.8626 (0.0002) & \textbf{-3.8628} (0.0000) & 0.077 & \textit{0.000} \\
        6 & hartmann6 & -3.0415 (0.0208) & -3.0419 (0.0292) & \textbf{-3.0425} (0.0000) & 0.703 & \textit{0.000} \\
        \hline
        5 & alpine01 & \textbf{0.0086} (0.0386) & 0.0235 (0.1771) & 0.0089 (0.0415) & \textit{0.028} & 0.598 \\
        10 & alpine01 & \textbf{0.2167} (0.1733) & 1.1542 (0.8509) & 0.6527 (0.8596) & \textit{0.000} & \textit{0.000} \\
        20 & alpine01 & \textbf{2.0290} (1.2094) & 6.3666 (2.8405) & 5.6045 (2.7298) & \textit{0.000} & \textit{0.000} \\
        \hline
        5 & styblinskiTang & -195.3928 (0.6391) & -195.2438 (0.7312) & \textbf{-195.7930} (0.1150) & 0.184 & \textit{0.000} \\
        10 & styblinskiTang & \textbf{-371.3045} (12.4763) & -364.309 (17.3369) & 359.7893 (11.3935) & \textit{0.011} & \textit{0.000} \\
        20 & styblinskiTang & \textbf{-720.4549 }(31.1087) & -679.9952 (55.7124) & -616.1962 (31.7896) & \textit{0.000} & \textit{0.000} \\
        \hline\hline
    \end{tabular}%
    }
    \caption{Best seen on collaborative/federated BO: median (standard deviation) on 30 independent runs: the lower the better (in bold, the lowest median value for each test problem).}
    \label{tab:2}
\end{table}

\begin{figure}[h!]
    \centering
    \includegraphics[width=0.32\linewidth]{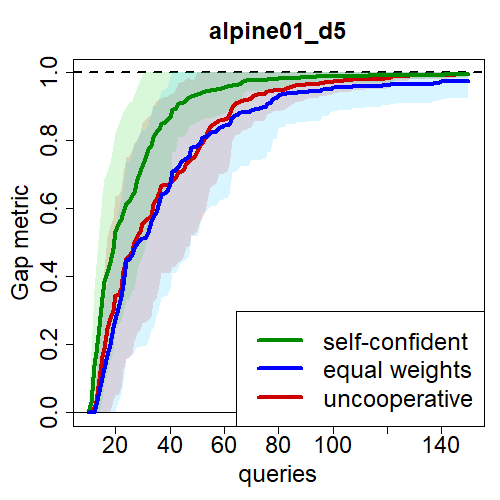}
    \includegraphics[width=0.32\linewidth]{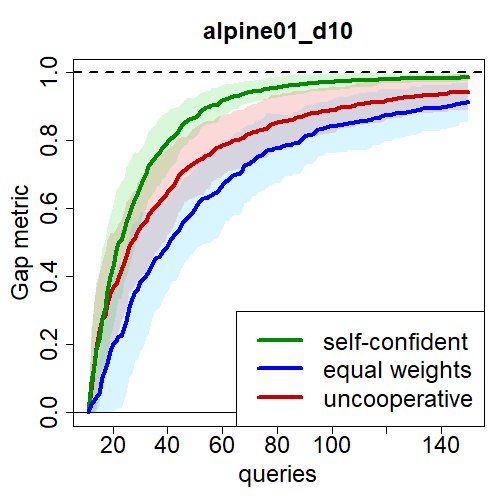}
    \includegraphics[width=0.32\linewidth]{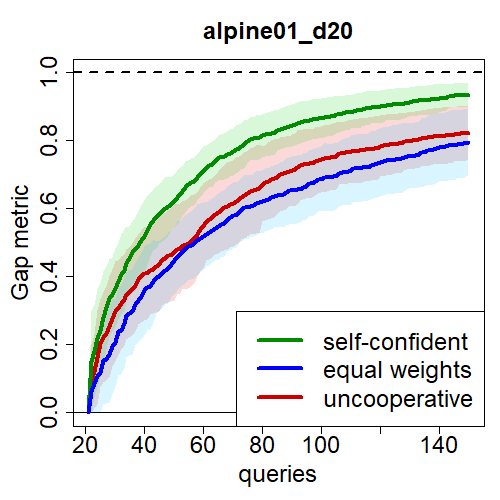}
    \caption{Collaborative/Federated BO: gap metric curves (median and standard deviation on 30 independent runs) of the three weighting scheme on the Alpine01 test problem and for different dimensionality of the search space ($d=\{5,10,20\}$.}
    \label{fig:5}
\end{figure}

\clearpage

\subsection{W2BGP for Batch BO: results}
In this section, we summarize the most relevant results on the application of W2BGP in batch BO.

\bmhead{Result \#3} \textit{The equal weighting schema (that is, simple model averaging) is the worst choice for W2BGP to target a (synchronous) batch BO task.}

Exactly as for collaborative/federated BO, the equal weighting schema (i.e., model averaging) applied to W2BGP leads to significantly poor performance also in the case of batch BO, in terms of both AUGC and best seen.

\bmhead{Result \#4} \textit{The uncooperative weighting schema is the most efficient choice for batch BO.}

As reported in Table \ref{tab:3}, the AUGC for the uncooperative schema is the largest for almost all the test problems considered, resulting in a higher sample efficiency compared to the other two weighting schemes. However, this seems not to be the case for problems with $d>=10$ (and also styblinskyTang with $d=5$), with differences in terms of AUGC that are relevant and statistically significant. Similar considerations apply to the best seen (Table \ref{tab:4}). The motivation behind the better performance of the uncooperative schema is quite trivial: the lack of cooperation -- i.e., knowledge sharing -- favors the generation of batch with well-diversified new queries, encouraging exploration. However, this bias towards exploration could become a drawback in high dimensions, where local and linear approaches have recently been reported as more effective and efficient \cite{hvarfner2024vanilla,eriksson2019scalable,doumont2025we}.

\begin{table}[h!]
    \centering
    \resizebox{\columnwidth}{!}{%
    \begin{tabular}{l|l|ccc|cc}
        \hline\hline        
        & \textbf{Test} & \textbf{self-confident} & \textbf{equally} & \textbf{uncooperative} & \textbf{U test} & \textbf{U test} \\
        d & \textbf{problem} & \textbf{W2BGPBO} & \textbf{W2BGPBO} & \textbf{parallel GPBO} & $p$\textbf{-value} & $p$\textbf{-value} \\
        & & \textcircled{a} & \textcircled{b} & \textcircled{c} & \textcircled{a} vs \textcircled{b} & \textcircled{a} vs \textcircled{c} \\
        \hline\hline
        1 & problem\_02 & \textbf{0.9266} (0.2322) & 0.8324 (0.2264) & 0.9183 (0.0393) & \textit{0.000} & 0.551 \\
        1 & problem\_03 & 0.8621 (0.1167) & 0.6310 (0.2735) & \textbf{0.8814} (0.1031) & \textit{0.000} & 0.579 \\
        1 & problem\_05 & 0.9217 (0.0466) & 0.7779 (0.1342) & \textbf{0.9272} (0.0437) & \textit{0.000} & 0.447 \\
        1 & problem\_07 & \textbf{0.9300} (0.0211) & 0.8655 (0.0653) & 0.9285 (0.0283) & \textit{0.000} & 0.742 \\
        1 & problem\_11 & 0.9444 (0.1380) & 0.8608 (0.1680) & \textbf{0.9483} (0.0340) & \textit{0.000} & 0.058 \\
        1 & problem\_14 & 0.8611 (0.0669) & 0.5865 (0.2356) & \textbf{0.8940} (0.0857) & \textit{0.000} & 0.393 \\
        1 & problem\_15 & 0.9370 (0.0938) & 0.8954 (0.2012) & \textbf{0.9443} (0.0464) & \textit{0.000} & \textit{0.031} \\
        1 & problem\_22 & 0.9330 (0.0283) & 0.8538 (0.1250) & \textbf{0.9439} (0.0397) & \textit{0.000} & 0.897 \\
        \hline
        2 & alpine01        & 0.9317 (0.0387) & 0.8231 (0.1709) & \textbf{0.9367} (0.0873) & \textit{0.000} & 0.188 \\
        2 & bird            & 0.9028 (0.0459) & 0.7522 (0.1809) & \textbf{0.9174} (0.0457) & \textit{0.000} & 0.188 \\
        2 & michalewicz     & 0.9140 (0.0449) & 0.7685 (0.1583) & \textbf{0.9184} (0.0741) & \textit{0.000} & 0.789 \\
        2 & styblinskiTang  & 0.9122 (0.0235) & 0.7343 (0.0896) & \textbf{0.9237} (0.0220) & \textit{0.000} & \textit{0.034} \\
        2 & ursem03         & \textbf{0.8553} (0.0686) & 0.5695 (0.2180) & 0.8403 (0.0843) & \textit{0.000} & 0.376 \\
        2 & ursemWaves      & \textbf{0.9552} (0.0251) & 0.8344 (0.1202) & 0.9646 (0.0344) & \textit{0.000} & 0.754 \\
        \hline
        3 & hartmann3       & 0.9612 (0.0610) & 0.8970 (0.0864) & \textbf{0.9618} (0.0177) & \textit{0.000} & 0.077 \\
        6 & hartmann6       & \textbf{0.7630} (0.0718) & 0.6888 (0.0803) & 0.7616 (0.0746) & \textit{0.000} & 0.092  \\
        \hline
        5 & alpine01 & \textbf{0.9270} (0.0491) & 0.8303 (0.1300) & 0.9079 (0.0670) & \textit{0.000} & \textit{0.008} \\
        10 & alpine01 & 0.8725 (0.0414) & 0.6879 (0.0906) & \textbf{0.8899} (0.0332) & \textit{0.000} & \textit{0.023} \\
        20 & alpine01 & 0.8246 (0.0438) & 0.6073 (0.0942) & \textbf{0.8394} (0.0317) & \textit{0.000} & \textit{0.002} \\
        \hline
        5 & styblinskiTang & \textbf{0.9225} (0.0213) & 0.7388 (0.0613) & 0.9144 (0.0200) & \textit{0.000} & 0.440 \\
        10 & styblinskiTang & \textbf{0.8124} (0.0520) & 0.5122 (0.1100) & 0.7273 (0.0475) & \textit{0.000} & \textit{0.000} \\
        20 & styblinskiTang & \textbf{0.6885} (0.0542) & 0.3851 (0.1061) & 0.5486 (0.0523) & \textit{0.000} & \textit{0.000} \\
        \hline\hline
    \end{tabular}%
    }
    \caption{AUGC on bacth BO: median (standard deviation) on 30 independent runs: the higher the better (in bold, the highest median value for each test problem).}
    \label{tab:3}
\end{table}

\begin{table}[h!]
    \centering
    \resizebox{\columnwidth}{!}{%
    \begin{tabular}{l|l|ccc|cc}
        \hline\hline        
        & \textbf{Test} & \textbf{self-confident} & \textbf{equally} & \textbf{uncooperative} & \textbf{U test} & \textbf{U test} \\
        d & \textbf{problem} & \textbf{W2BGPBO} & \textbf{W2BGPBO} & \textbf{parallel GPBO} & $p$\textbf{-value} & $p$\textbf{-value} \\
        & & \textcircled{a} & \textcircled{b} & \textcircled{c} & \textcircled{a} vs \textcircled{b} & \textcircled{a} vs \textcircled{c} \\
        \hline\hline
        
        1 & problem\_02 &  -1.8996 (0.1775) &  -1.8996 (0.1775) &  \textbf{-1.8996 (0.0000)} & 0.053 & 0.346 \\
        1 & problem\_03 & -12.0312 (0.4827) & -11.6719 (2.3210) & \textbf{-12.0312 (0.0015)} & \textit{0.000} & 0.689 \\
        1 & problem\_05 &  -1.4891 (0.0007) &  -1.4890 (0.0139) &  \textbf{-1.4891 (0.0000)} & \textit{0.000} & 1.000 \\
        1 & problem\_07 &  \textbf{-1.6013 (0.0000)} &  -1.6013 (0.0001) &  \textbf{-1.6013 (0.0000)} & \textit{0.015} & 1.000 \\
        1 & problem\_11 &  \textbf{-1.5000 (0.0000)} &  -1.4999 (0.0005) &  \textbf{-1.5000 (0.0000)} & \textit{0.000} & 1.000 \\
        1 & problem\_14 &  \textbf{-0.7887 (0.0000)} &  -0.7886 (0.1829) &  \textbf{-0.7887 (0.0000)} & \textit{0.001} & 1.000 \\
        1 & problem\_15 &  -0.0355 (0.0039) &  -0.0355 (0.0076) &  \textbf{-0.0355 (0.0000)} & \textit{0.014} & \textit{0.022} \\
        1 & problem\_22 &  \textbf{-1.0000 (0.0000)} &  -0.9999 (0.0017) &  \textbf{-1.0000 (0.0000)} & \textit{0.000} & 1.000 \\
        \hline
        2 & alpine01 &    \textbf{0.0018} (0.0010) &    0.0026 (0.0636)  &    0.0022 (0.0010) & \textit{0.003} & 0.447 \\
        2 & bird            & -106.7642 (0.0003) & -106.7564 (19.7754) & \textbf{-106.7645} (0.0000) & \textit{0.000} & \textit{0.000} \\
        2 & michalewicz     &   \textbf{-1.8013 (0.0000)} &   -1.8012 (0.0051)  &   \textbf{-1.8013 (0.0000)} & \textit{0.000} & 1.000 \\
        2 & styblinskiTang  &  -78.3319 (0.0006) &  -78.3038 (0.1275)  &  \textbf{-78.3323} (0.0000) & \textit{0.000} & \textit{0.000} \\
        2 & ursem03         &   -2.9993 (0.0006) &   -2.9549 (0.4688)  &   \textbf{-2.9996} (0.0011) & \textit{0.000} & 0.914 \\
        2 & ursemWaves      &   -7.3069 (0.0001) &   -7.3052 (0.0012)  &   \textbf{-7.3070} (0.0000) & \textit{0.000} & \textit{0.000} \\
        \hline
        3 & hartmann3       &   \textbf{-3.8628 (0.0000)} &   -3.8627 (0.0012)  &   \textbf{-3.8628 (0.0000)} & \textit{0.000} & 0.072 \\
        6 & hartmann6       &   -3.0424 (0.0000) & -3.0423 (0.0275) & \textbf{-3.0425} (0.0000) & \textit{0.000} & \textit{0.000} \\
        \hline
        5 & alpine01 & \textbf{0.0639} (0.0503) & 0.1009 (0.2647) & 0.0982 (0.0785) & \textit{0.028} & \textit{0.001} \\
        10 & alpine01 & 0.4873 (0.3509) & 1.5070 (1.1214) & \textbf{0.3685} (0.1822) & \textit{0.000} & \textit{0.011} \\
        20 & alpine01 & 0.9622 (0.7231) & 6.4980 (3.4238) & \textbf{0.7164} (0.2757) & \textit{0.000} & \textit{0.008} \\
        \hline
        5 & styblinskiTang & -195.8093 (0.0136) & -195.1673 (2.7078) & \textbf{-195.8308} (0.0001) & \textit{0.000} & \textit{0.000} \\
        10 & styblinskiTang & \textbf{-387.1686} (7.5783) & -354.9120 (16.6438) & 374.8057 (6.7772) & \textit{0.000} & \textit{0.001} \\
        20 & styblinskiTang & \textbf{-714.1044} (25.009) & -627.7947 (53.8764) & 661.2477 (17.8840) & \textit{0.000} & \textit{0.000}  \\
        \hline\hline
    \end{tabular}%
    }
    \caption{Best seen on batch BO: median (standard deviation) on 30 independent runs: the lower the better (in bold, the lowest median value for each test problem).}
    \label{tab:4}
\end{table}

\clearpage

\subsection{WBGP for multi-fidelity BO: results}
Finally, we summarize here the most relevant results on the application of W2BGP for multi-fidelity BO.

\bmhead{Result \#5} \textit{Using fidelities directly as weights leads to the worst AUGC values.}

As reported in Table \ref{tab:5}, a simple model averaging schema of $M$ GPs with weights equal to their own fidelities resulted in the lowest AUGC values, leading to an inefficient MFBO implementation. On the other hand, in some cases, it converged to better values of the best seen (Table \ref{tab:6}).

\bmhead{Result \#6} \textit{An equal-weights schema could be the best choice}.

According to Table \ref{tab:5}, there is no clear winner between \textit{rescaled-weights} and \textit{equal-weights} schemes, because each one of them provided the largest value of AUGC in exactly half of the test problems. However, the difference between AUGC values is quite small when the winner is the \textit{rescaled-weights} schema, while it is relevant in the opposite cases. Thus, the \textit{equal-weights} schema appears as the most reasonable choice.
\begin{table}[h!]
    \centering
    \resizebox{\columnwidth}{!}{%
    \begin{tabular}{l|l|l|ccc|cc}
        \hline\hline        
        & & \textbf{Test} & \textbf{fidelities} & \textbf{rescaled} & \textbf{equal} & \textbf{U test} & \textbf{U test} \\
        d & $n_s$ & \textbf{problem} & \textbf{as weights} & \textbf{weights} & \textbf{weights} & $p$\textbf{-value} & $p$\textbf{-value} \\
        & & & \textcircled{a} & \textcircled{b} & \textcircled{c} & \textcircled{a} vs \textcircled{c} & \textcircled{b} vs \textcircled{c} \\
        \hline\hline
        1 & 4 & forrester & 4.6773 (10.159)	& 5.2236 (10.6378) & \textbf{11.3334} (9.2998) & 0.233 & 1.000 \\
        \hline
        2 & 3 & rosenbrock & 37.011 (15.2673) & \textbf{39.3278} (15.0407) & 33.734 (13.2947) & 0.839 & 1.000 \\
        5 & 3 & rosenbrock &  78.4977 (38.6431) & 75.8972 (30.1297)	& \textbf{79.6312} (31.2393) & 0.318 & 1.000 \\
        \hline
        2 & 3 & sr-rastrigin & 1.3092 (13.0303)	& 7.1723 (14.4443) & \textbf{12.9688} (16.3295) & \textit{0.018} & \textit{0.000} \\
        \hline
        1 & 2 & heterogeneous & 12.1809 (11.2593) & \textbf{18.3367} (9.7259) & 17.9024 (10.0064) & 0.360 & 1.000 \\
        2 & 2 & heterogeneous & 38.4363 (18.0917) & 40.3747 (18.8787) & \textbf{50.0000} (13.8076) & \textit{0.010} & \textit{0.000} \\
        3 & 2 & heterogeneous & 76.1669 (20.9419) & \textbf{80.0000} (15.5379) & 79.0463 (20.8562) & 1.000 & \textit{0.000} \\
        \hline
        2 & 2 & pacioreck & 50.6007 (15.1277) & \textbf{51.3141} (17.2077) & 50.3113 (42.1495) & 0.895 & \textit{0.000} \\
        \hline\hline
    \end{tabular}%
    }
    \caption{AUGC on multi-fidelity BO: median (standard deviation) on 30 independent runs: the higher the better (in bold, the highest median value for each test problem).}
    \label{tab:5}
\end{table}

\vspace{-1.25cm}

\begin{table}[h!]
    \centering
    \resizebox{\columnwidth}{!}{%
    \begin{tabular}{l|l|l|ccc|cc}
        \hline\hline        
        & & \textbf{Test} & \textbf{fidelities} & \textbf{rescaled} & \textbf{equal} & \textbf{U test} & \textbf{U test} \\
        d & $n_s$ & \textbf{problem} & \textbf{as weights} & \textbf{weights} & \textbf{weights} & $p$\textbf{-value} & $p$\textbf{-value} \\
        & & & \textcircled{a} & \textcircled{b} & \textcircled{c} & \textcircled{a} vs \textcircled{c} & \textcircled{b} vs \textcircled{c} \\
        \hline\hline
        1 & 4 & forrester & \textbf{-5.5217} (1.9288) & -5.3427 (2.0728) & -5.3756 (1.4054) & 0.446 & \textit{0.000} \\
        \hline
        2 & 3 & rosenbrock & \textbf{1.2014} (1.7257) & 1.9420 (1.3894) & 1.6995 (3.9419) & \textit{0.045} & \textit{0.000} \\
        5 & 3 & rosenbrock & \textbf{32.2673} (74.7717) & 43.2524 (59.7992) & 33.5468 (47.0825) & 0.952 & 1.000 \\
        \hline
        2 & 3 & sr-rastrigin & 0.8756 (0.4935) & 0.5773 (0.4838) & \textbf{0.1778 }(0.4996) & \textit{0.006} & \textit{0.000} \\
        \hline
        1 & 2 & heterogeneous & -0.6058 (0.1658) & \textbf{-0.6182 (0.1235)} & -0.6182 (0.1240) & 0.324 & \textit{0.000} \\
        2 & 2 & heterogeneous & -0.5627 (0.1042) & -0.5627 (0.0923) & \textbf{-0.5627 (0.0477)} & 0.050 & \textit{0.000} \\
        3 & 2 & heterogeneous & -0.5627 (0.0511) & \textbf{-0.5627 (0.0289)} & -0.5627 (0.0436) & 0.590 & 1.000 \\
        \hline
        2 & 2 & pacioreck & -0.9968 (0.1294) & -0.9995 (0.1002) & \textbf{-0.9989} (0.0129) & 0.140 & \textit{0.000} \\
        \hline\hline
    \end{tabular}%
    }
    \caption{Best seen on multi-fidelity BO: median (standard deviation) on 30 independent runs: the lower the better (in bold, the lowest median value for each test problem).}
    \label{tab:6}
\end{table}

\vspace{-0.75cm}

According to Result \#6, it seems that the most relevant contribution is given by the acquisition function instead of the specific weighting schema. This is reasonable because the penalization term in the denominator of the acquisition function (\ref{eq:mfbo_acq}) -- i.e., the Wasserstein distance between two univariate GDs -- allows us to account for the distance between the W2BGP and each source-specific GP, regardless of the weighting schema used in the W2BGP. This is also confirmed by the almost equal usage of the ground-truth made by the three schemes (Table \ref{tab:7}).

\begin{table}[h!]
    \centering
    \begin{tabular}{l|l|l|ccc}
        \hline\hline        
        & & \textbf{Test} & \textbf{fidelities} & \textbf{rescaled} & \textbf{equal} \\
        d & $n_s$ & \textbf{problem} & \textbf{as weights} & \textbf{weights} & \textbf{weights}\\
        \hline\hline
        1 & 4 & forrester & 60.56\% & 62.96\% & 60.09\% \\
        \hline
        2 & 3 & rosenbrock & 50.39\% & 50.54\% & 51.81\% \\
        5 & 3 & rosenbrock & 50.08\% & 50.04\% & 50.08\% \\
        \hline
        2 & 3 & sr-rastrigin & 55.78\% & 56.76\% & 64.46\% \\
        \hline
        1 & 2 & heterogeneous & 62.50\% & 60.00\% & 62.40\% \\
        2 & 2 & heterogeneous & 56.82\% & 58.70\% & 58.80\% \\
        3 & 2 & heterogeneous & 50.45\% & 50.28\% & 50.21\% \\
        \hline
        2 & 2 & pacioreck & 84.53\% & 82.86\% & 86.20\% \\
        \hline\hline
    \end{tabular}
    \caption{Usage (in percentage) of the ground-truth information source in multi-fidelity BO.}
    \label{tab:7}
\end{table}

\section{Conclusions}
As proposed and empirically evaluated in this paper, the weighted Wasserstein Barycenter of Gaussian Process (W2BGP) can represent a unifying framework for exotic BO tasks which requires the combination of many GPs into a unique model.

Our empirical analysis demonstrate that a suitable weighting schema can be reasonably identified depending on the target BO task, at least among the three considered in the paper: collaborative/federated BO, synchronous batch-BO, and multi-fidelity BO. As future work, our research is going to address other exotic BO tasks, such as transfer-BO (also known as transfer-learning for BO), scalable and sparse GPs for BO, as well as global-local BO.

Another relevant research direction refers to the settings in which a reasonable weighting schema cannot be identified depending on the task, or an optimal weighting schema is desired with respect to some scoring function. An adaptive weighting schema could be considered in these cases, similarly to the one recently proposed in \cite{ziomek2024time} -- event if, their approach is delayed by one iteration because the last function evaluation is used to calculate the prediction errors of the GPs, then used as weights for combining them into a unique model.

Finally, another important result of our study is the interpretation of the most well-known BO acquisition function under the lens of the proposed W2BGP, enabled by the analogy between GDs and the GP's posterior. The analysis and interpretation of other BO acquisition functions could be another interesting research perspective.\\

\bmhead{Data and code availability} Both the code and the data supporting the findings of this study are openly available at the following Github repository:\\
\url{https://github.com/acandelieri/W2BGP\_for\_exotic\_BO.git}

\clearpage

\begin{appendices}

\section{Gap metrics charts}\label{secA}

\subsection{Gap metrics charts for collaborative/federated BO}

\begin{figure}[H]
    \centering
    \includegraphics[width=0.32\linewidth]{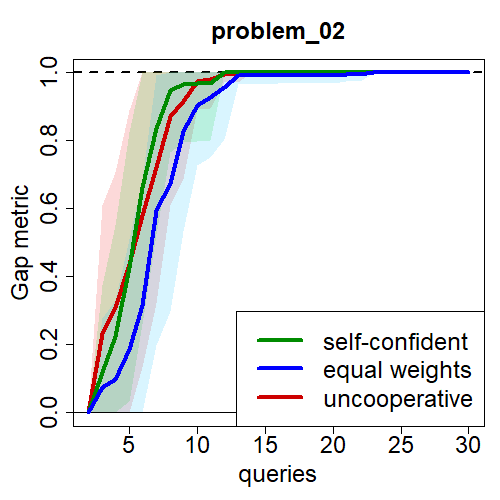}
    \includegraphics[width=0.32\linewidth]{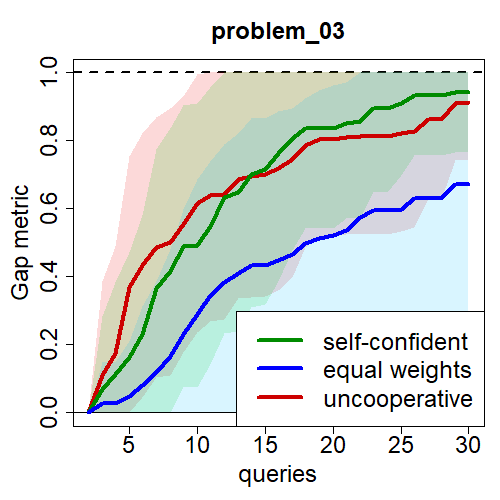}
    \includegraphics[width=0.32\linewidth]{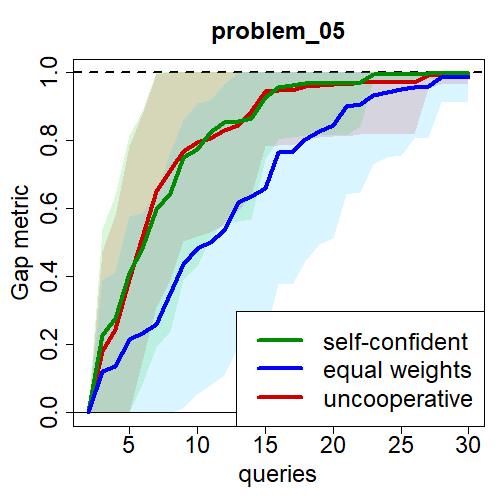}
    \includegraphics[width=0.32\linewidth]{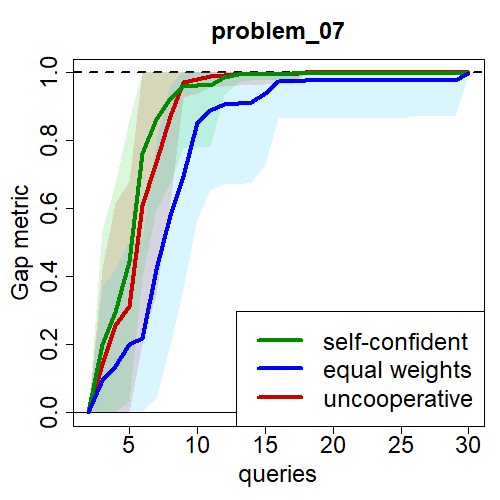}
    \includegraphics[width=0.32\linewidth]{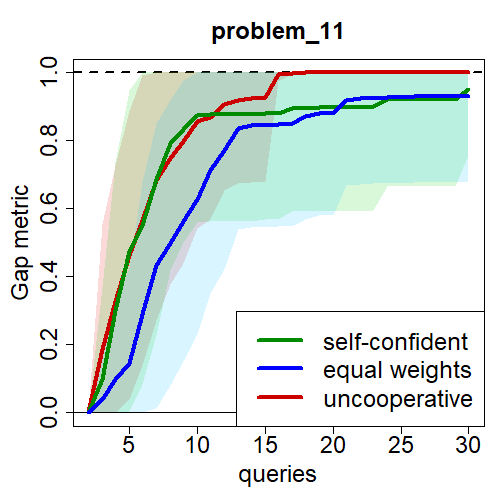}
    \includegraphics[width=0.32\linewidth]{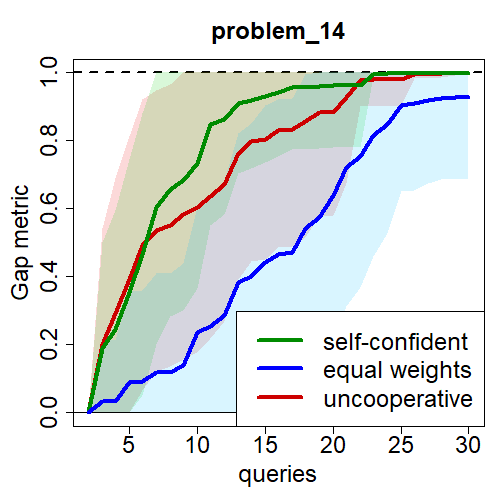}
    \includegraphics[width=0.32\linewidth]{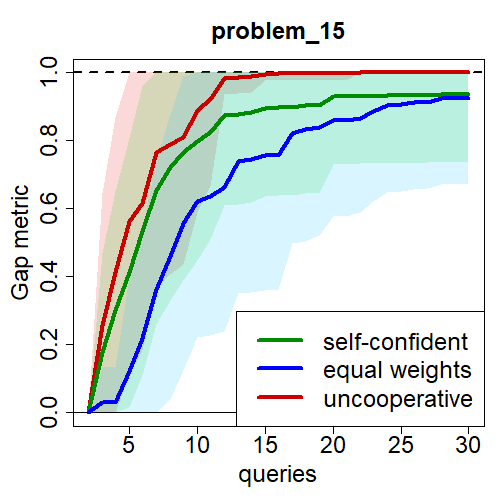}
    \includegraphics[width=0.32\linewidth]{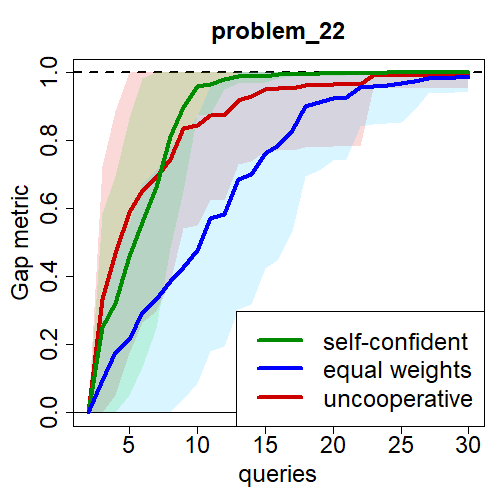}
    \caption{Collaborative/Federated BO: gap metric curves for the eight 1-dimensional test problems addressed.}
    \label{fig:apx1-a}
\end{figure}

\newpage
\begin{figure}[H]
    \centering
    \includegraphics[width=0.24\linewidth]{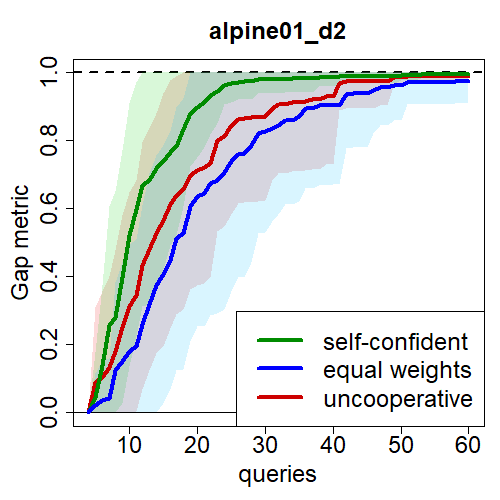}
    \includegraphics[width=0.24\linewidth]{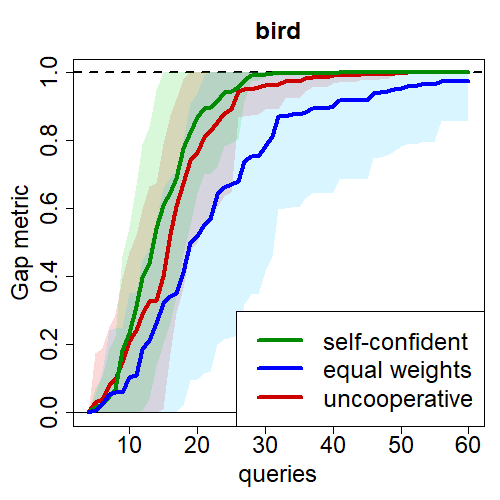}
    \includegraphics[width=0.24\linewidth]{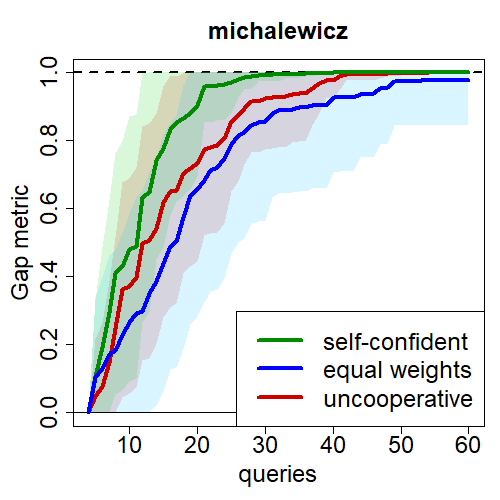}
    \includegraphics[width=0.24\linewidth]{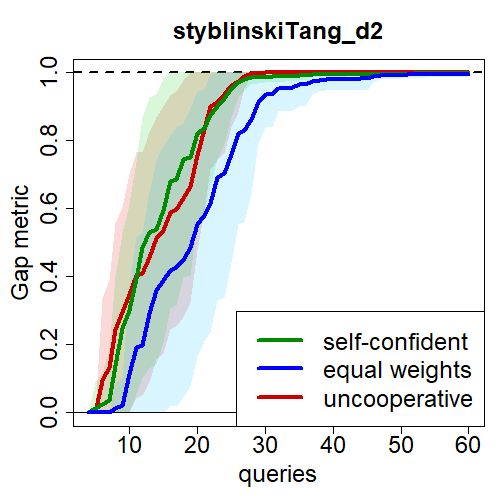}
    \includegraphics[width=0.24\linewidth]{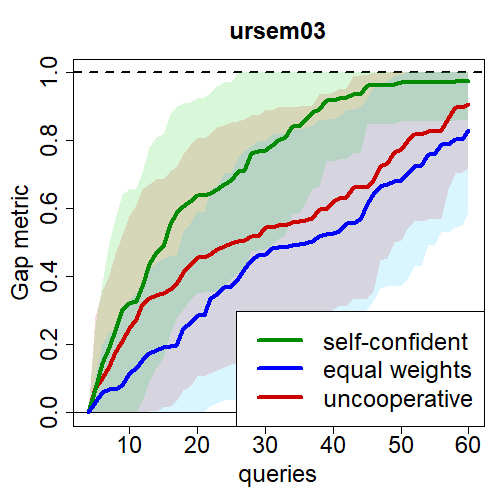}
    \includegraphics[width=0.24\linewidth]{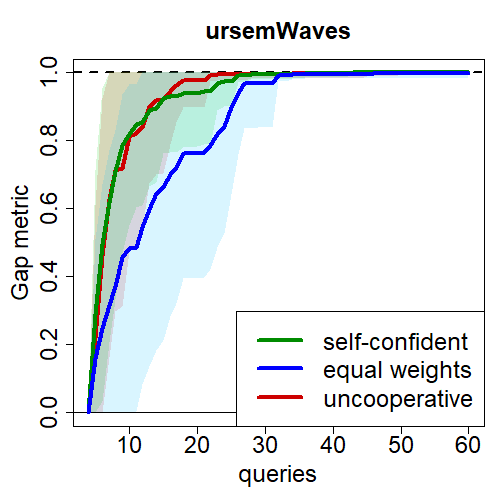}
    \includegraphics[width=0.24\linewidth]{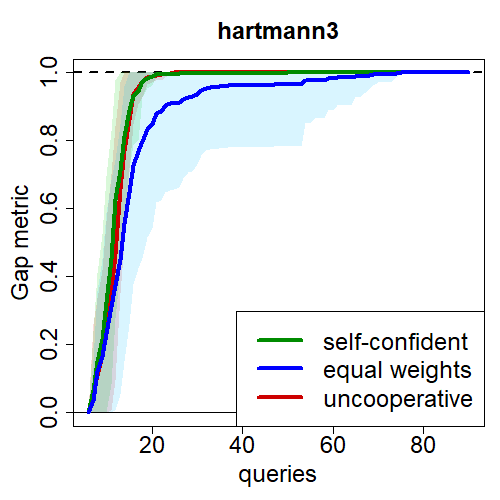}
    \includegraphics[width=0.24\linewidth]{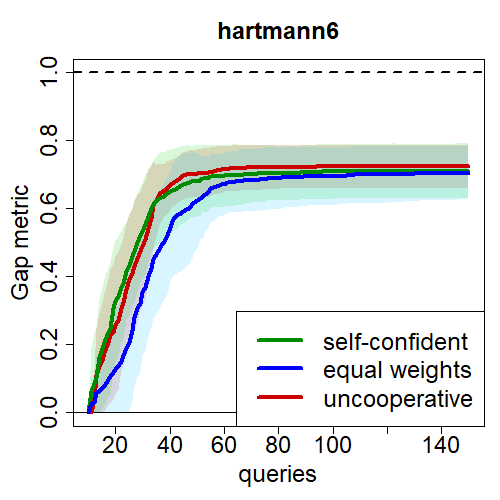}
    \includegraphics[width=0.24\linewidth]{appendix_figures/gap_metrics_federatedBO/GapMetric_federatedBO_18.png}
    \includegraphics[width=0.24\linewidth]{appendix_figures/gap_metrics_federatedBO/GapMetric_federatedBO_19.png}
    \includegraphics[width=0.24\linewidth]{appendix_figures/gap_metrics_federatedBO/GapMetric_federatedBO_20.png}
    \includegraphics[width=0.24\linewidth]{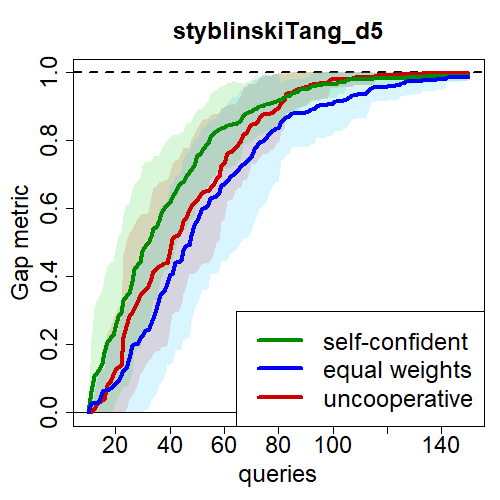}
    \includegraphics[width=0.24\linewidth]{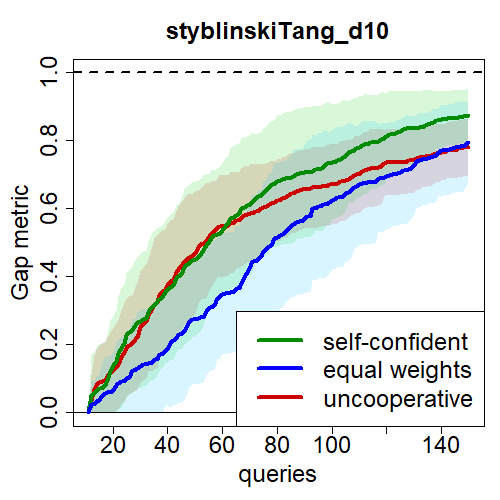}
    \includegraphics[width=0.24\linewidth]{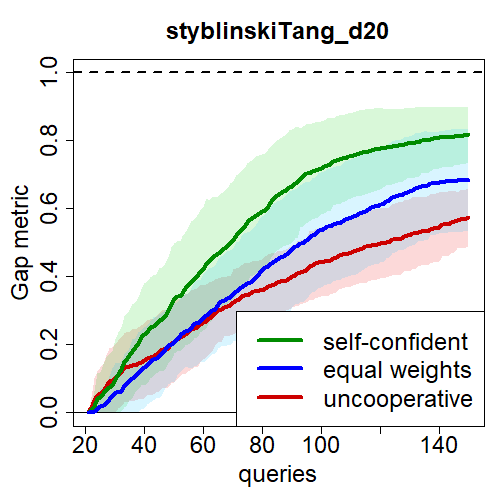}    
    \caption{Collaborative/Federated BO: gap metric curves for the fourteen $d$-dimensional test problems addressed, with $d\geq2$.}
    \label{fig:apx1-b}
\end{figure}

\clearpage
\subsection{Gap metrics charts for batch BO}

\begin{figure}[H]
    \centering
    \includegraphics[width=0.32\linewidth]{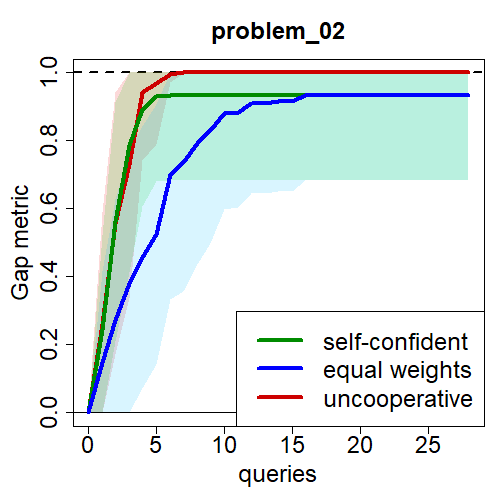}
    \includegraphics[width=0.32\linewidth]{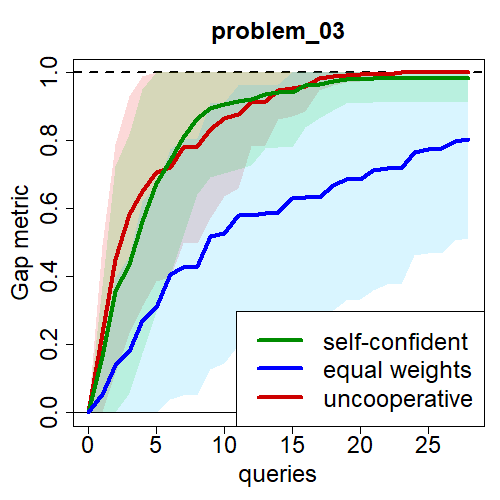}
    \includegraphics[width=0.32\linewidth]{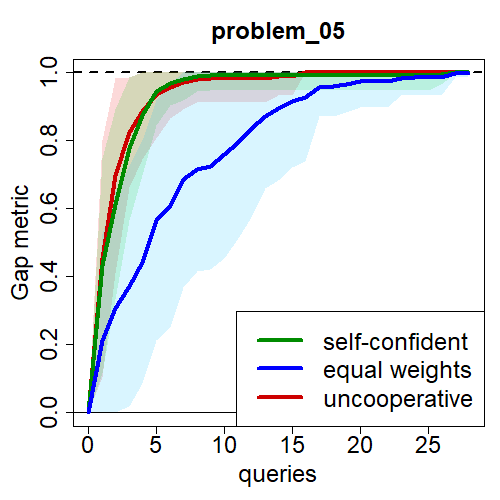}
    \includegraphics[width=0.32\linewidth]{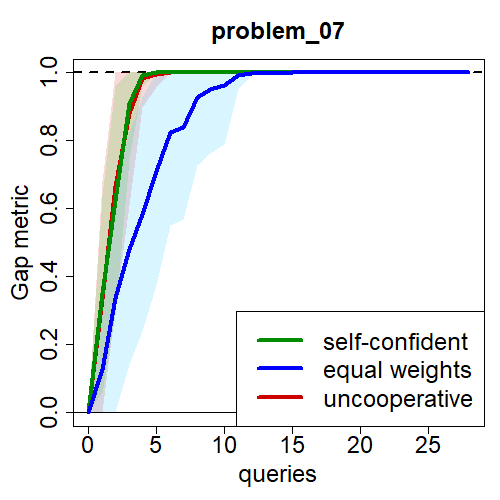}
    \includegraphics[width=0.32\linewidth]{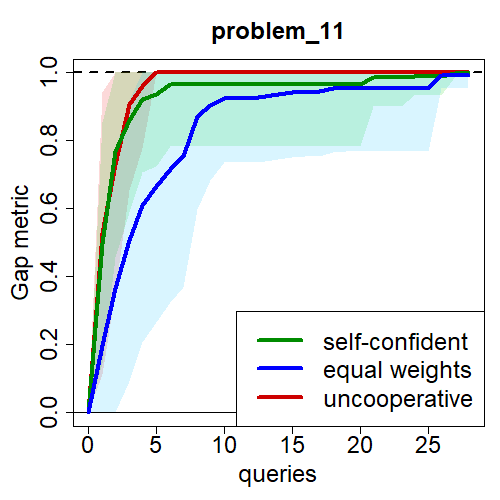}
    \includegraphics[width=0.32\linewidth]{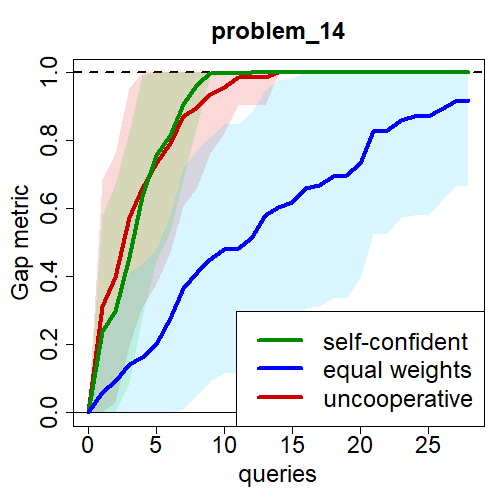}
    \includegraphics[width=0.32\linewidth]{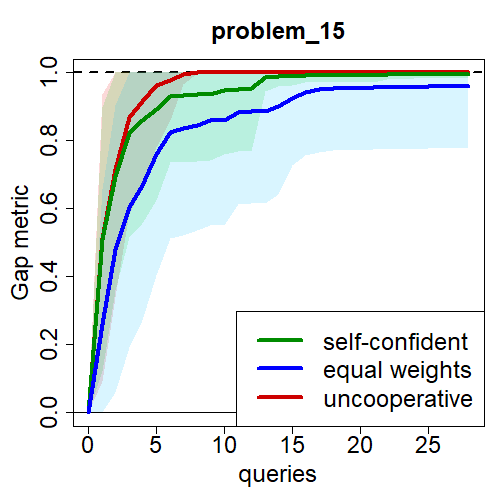}
    \includegraphics[width=0.32\linewidth]{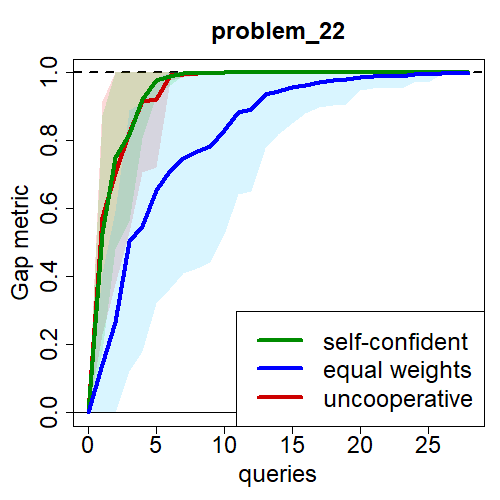}
    \caption{Batch BO: gap metric curves for the eight 1-dimensional test problems addressed.}
    \label{fig:apx2-a}
\end{figure}

\newpage
\begin{figure}[H]
    \centering
    \includegraphics[width=0.24\linewidth]{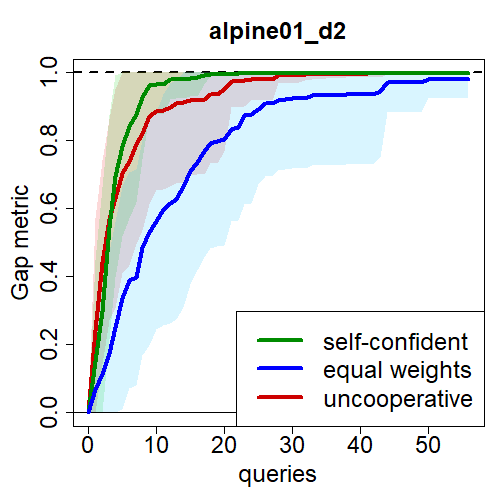}
    \includegraphics[width=0.24\linewidth]{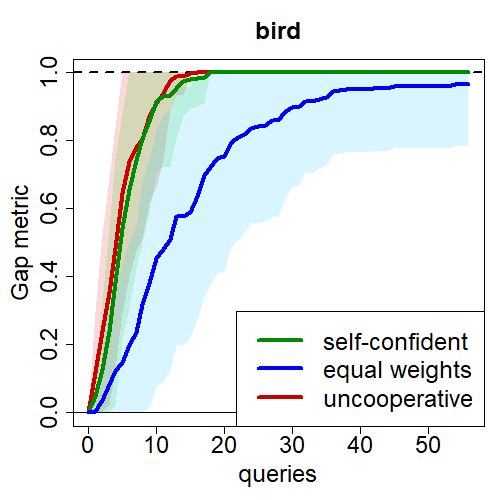}
    \includegraphics[width=0.24\linewidth]{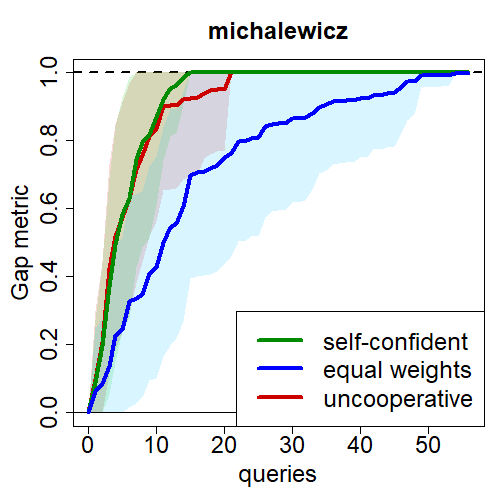}
    \includegraphics[width=0.24\linewidth]{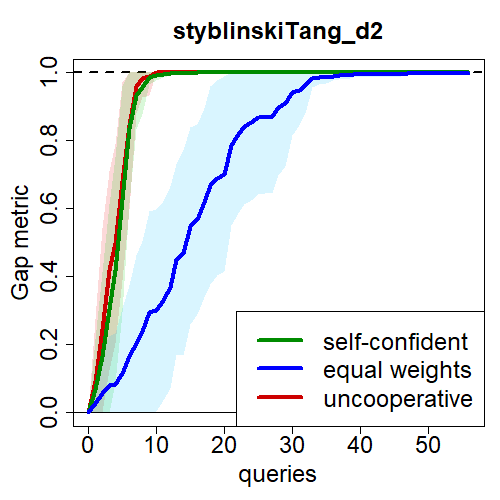}
    \includegraphics[width=0.24\linewidth]{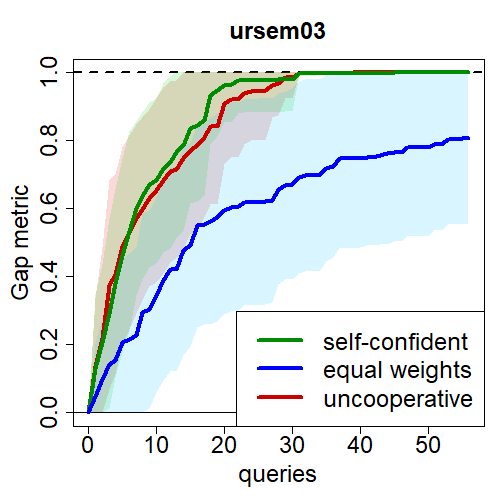}
    \includegraphics[width=0.24\linewidth]{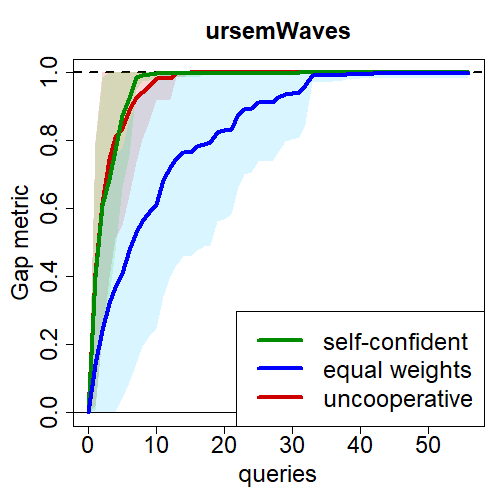}
    \includegraphics[width=0.24\linewidth]{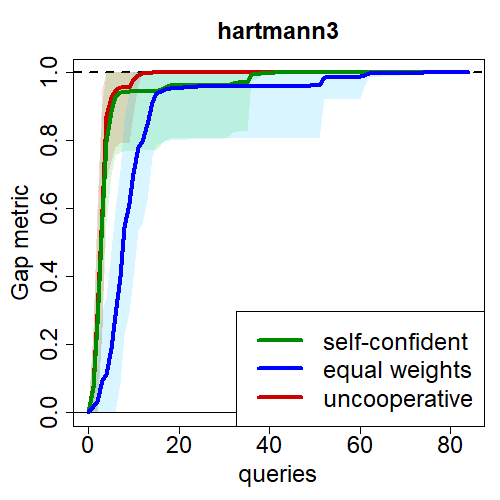}
    \includegraphics[width=0.24\linewidth]{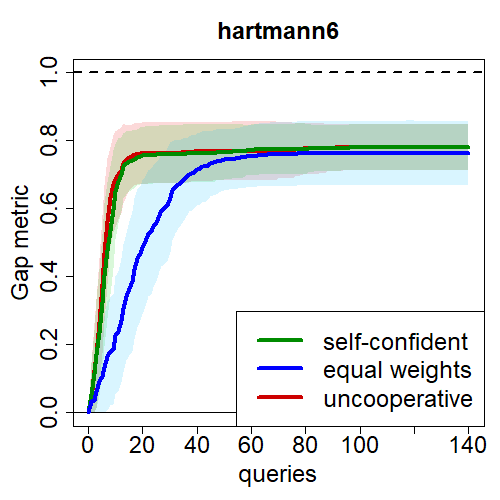}
    \includegraphics[width=0.24\linewidth]{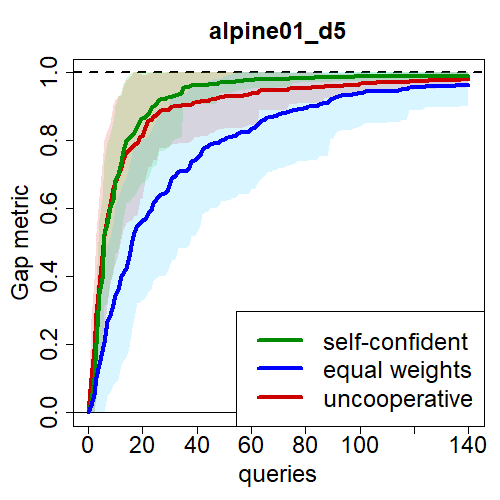}
    \includegraphics[width=0.24\linewidth]{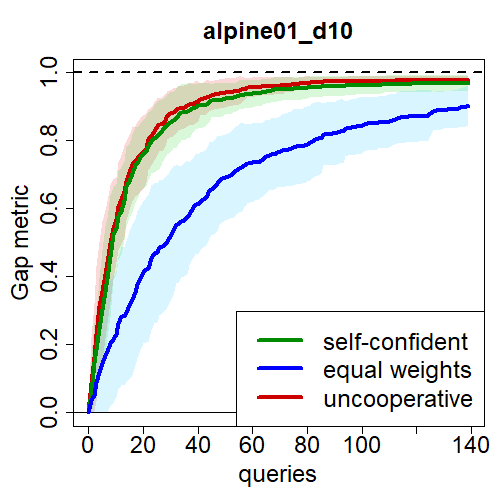}
    \includegraphics[width=0.24\linewidth]{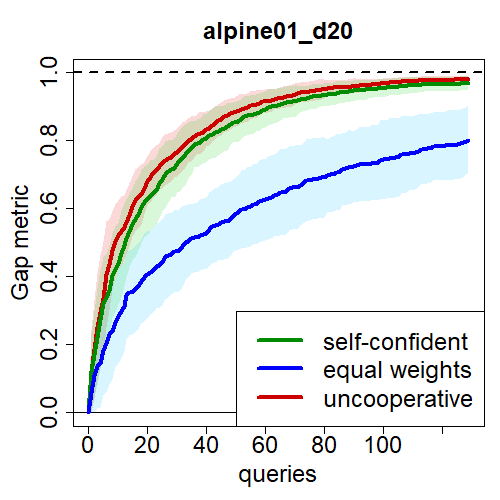}
    \includegraphics[width=0.24\linewidth]{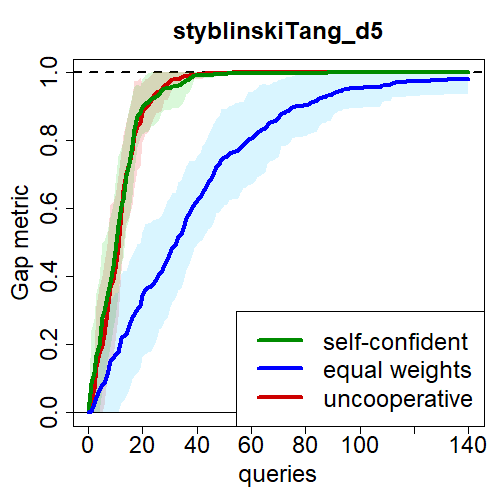}
    \includegraphics[width=0.24\linewidth]{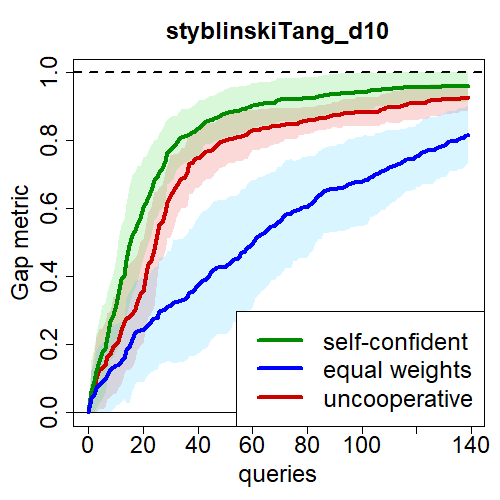}
    \includegraphics[width=0.24\linewidth]{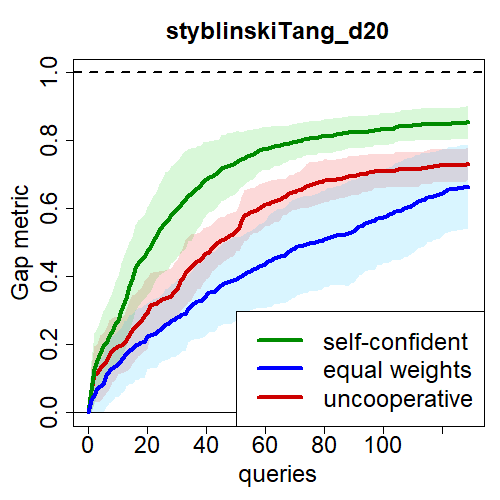}
    \caption{Batch BO: gap metric curves for the fourteen $d$-dimensional test problems addressed, with $d\geq2$.}
    \label{fig:apx2-b}
\end{figure}

\clearpage
\subsection{Gap metrics charts for multifidelity BO}

\begin{figure}[H]
    \centering
    \includegraphics[width=0.32\linewidth]{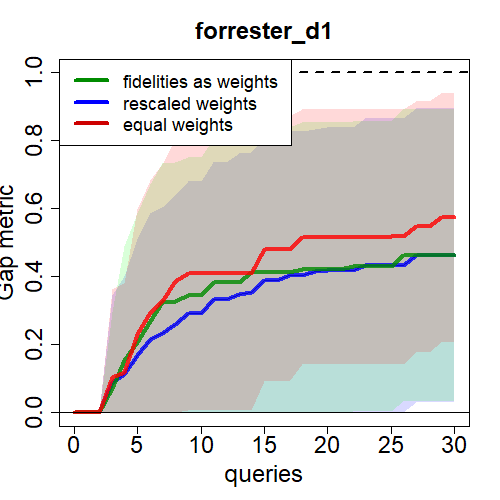}
    \includegraphics[width=0.32\linewidth]{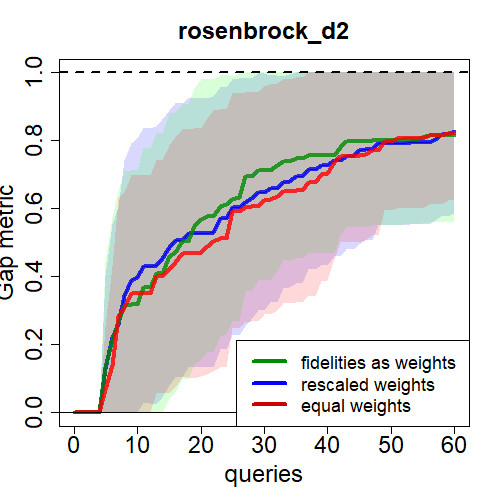}
    \includegraphics[width=0.32\linewidth]{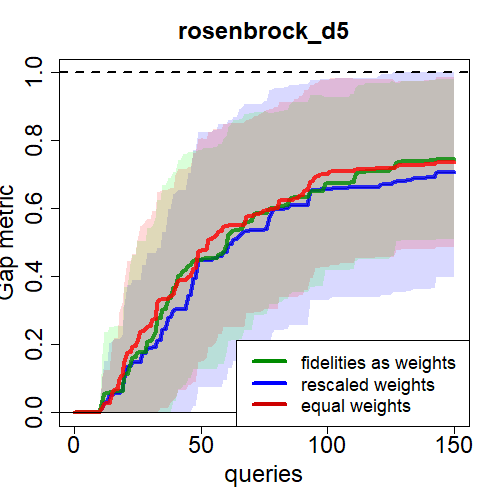}
    \includegraphics[width=0.32\linewidth]{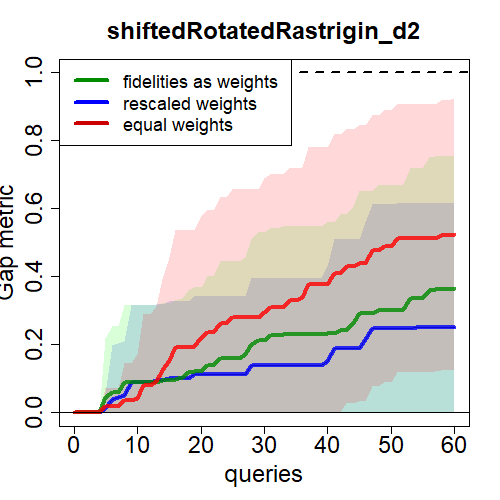}
    \includegraphics[width=0.32\linewidth]{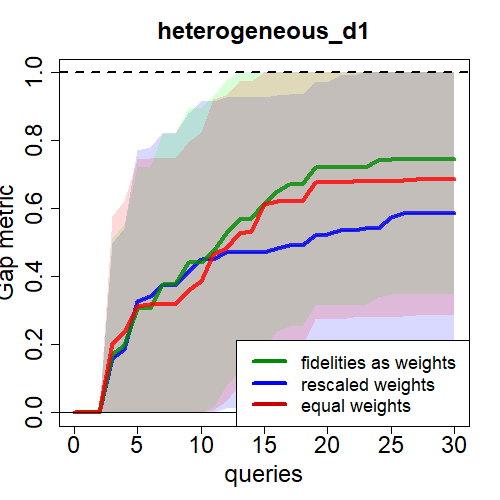}
    \includegraphics[width=0.32\linewidth]{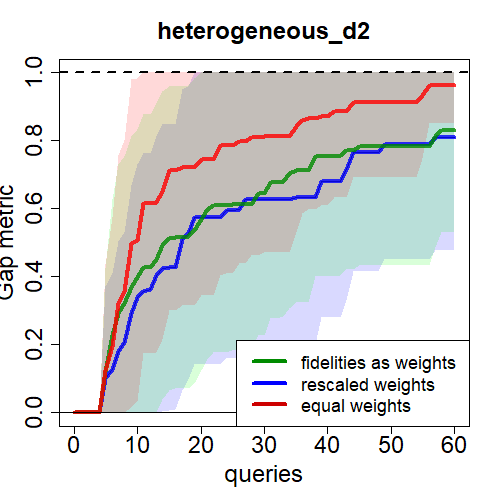}
    \includegraphics[width=0.32\linewidth]{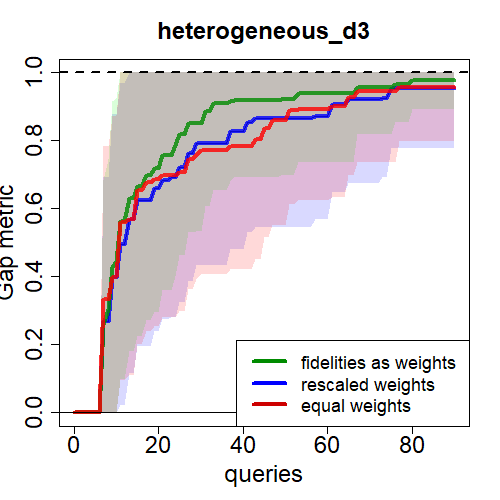}
    \includegraphics[width=0.32\linewidth]{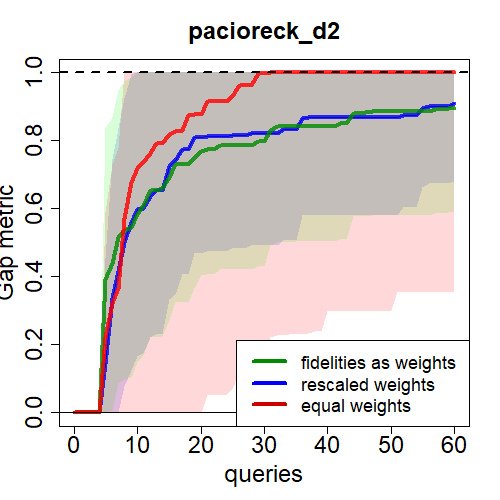}
    \caption{Multi-fidelity: gap metric curves for the test problems addressed.}
    \label{fig:apx3}
\end{figure}

\clearpage
\section{Test problems}\label{secB}

\subsection{Test problems for collaborative/federated and batch BO}

\begin{table}[h!]
    \centering
    \resizebox{\columnwidth}{!}{%
    \begin{tabular}{|l|c|c|c|c|}
         \hline
         Test problem & $d$ & search space & equation & $f(x^*)$ \\
         \hline\hline
         problem 02 & 1 & $[2.7,7.5]$ & $f(x)=\sin(x)\sin\Big(\frac{10x}{3}\Big)$ & -1.8996\\
         problem 03 & 1 & $[-10,10]$ & $f(x)= -\sum_{k=0}^5 \Big[ k \sin\big( (k+1)x + k\big) \Big]$ & -12.0312\\
         problem 05 & 1 & $[0, 1.2]$ & $f(x)=-(1.4 - 3x)\sin(18x)$ & -1.4891\\
         problem 07 & 1 & $[2.7,7.5]$ & $f(x) = \sin(x) + \sin\Big(\frac{10x}{3}\Big) + \log(x) - 0.84x + 3$ & -1.6013\\
         problem 11 & 1 & $[-\pi,2\pi]$ & $f(x) = 2\cos(x) + \cos(2x)$ & -1.5000\\
         problem 14 & 1 & $[0,4]$ & $f(x) = -e^{-x} \sin(2\pi x)$ & -0.7887\\
         problem 15 & 1 & $[-5,5]$ & $f(x) = \frac{x^2 - 5x + 6}{x^2 + 1}$ & -0.03553391\\
         problem 22 & 1 & $[0,20]$ & $f(x) = e^{-3x} - \big(\sin(x)\big)^3$ & $\exp(-\frac{27\pi}{2})-1$\\     
         \hline
         alpine01 & any & $[-10,10]^d$ & $ f(x) = \sum_{i=1}^d \Big\vert x \sin(x) + 0.1x\Big\vert$ & 0.0 \\
         bird & 2 & $[-2\pi,2\pi]^2$ & $f(x) = (x_1-x_2)^2 + e^{ (1-\sin(x_1))^2 } \cos(x_2) + e^{ (1-\cos(x_2))^2 } \sin(x_1)$ & -106.7645\\
         hartmann3 & 3 & $[0,1]^3$ & $f(x) = -\sum_{i=1}^4 \alpha_i \exp\big[-\sum_{j=1}^3 A_{ij (x_j-P_{ij})^2}\big]$ & -3.862782145\\ 
         & & & where $\alpha, \textbf{A}$, and $\textbf{P}$ as in \url{https://www.sfu.ca/~ssurjano/hart3.html} & \\
         hartmann6 & 6 & $[0,1]^6$ & $f(x) = -\sum_{i=1}^4 \alpha_i \exp\big[-\sum_{j=1}^3 A_{ij (x_j-P_{ij})^2}\big]$ & -3.32237\\
         & & & where $\alpha, \textbf{A}$ and $\textbf{P}$ as in \url{https://www.sfu.ca/~ssurjano/hart6.html}  & \\
         michalewicz & 2 & $[0,\pi]^2$ & $f(x) = -\sum_{i=1}^d \sin(x) \Big(\sin\big((i x^2\big)/\pi)\Big)^{2m}$, with $m=10$ & -1.8013\\
         styblinskiTang & any & $[-5,5]^d$ & $f(x) = \frac{1}{2} \sum_{i=1}^d \Big(x_i^4-16x_i^2+5x_i\Big)$ & $-39.16599\cdot d$\\
         ursem03 & 2 & $[-2,2]\times[-1.5,1.5]$ & $f(x) =  -\sin(2.2\pi x_1+0.5\pi) \frac{2-\vert x_1\vert}{2} \frac{3-\vert x_1\vert)}{2} - \sin(2.2 \pi x_2+0.5\pi) \frac{2-\vert x_2\vert}{2} \frac{3-\vert x_2\vert}{2}$ & -3.0000\\
         ursemWaves & 2 & $[-0.9,1.2]\times[-1.2,1.2]$ & $f(x) = -(0.3 x_1)^3 + (x_2^2 - 4.5 x_2^2) x_1x_2 + 4.7 \cos(3x_1 - x_2^2 (2+x_1)) \sin(2.5 \pi x_1)$ & -7.306999\\
         \hline
    \end{tabular}%
    }
    \caption{Definition of all the test problems considered in the paper, for both Collaborative/Federated BO and Batch BO.}
\end{table}

\clearpage

\begin{figure}[h!]
    \centering
    \includegraphics[width=0.95\linewidth]{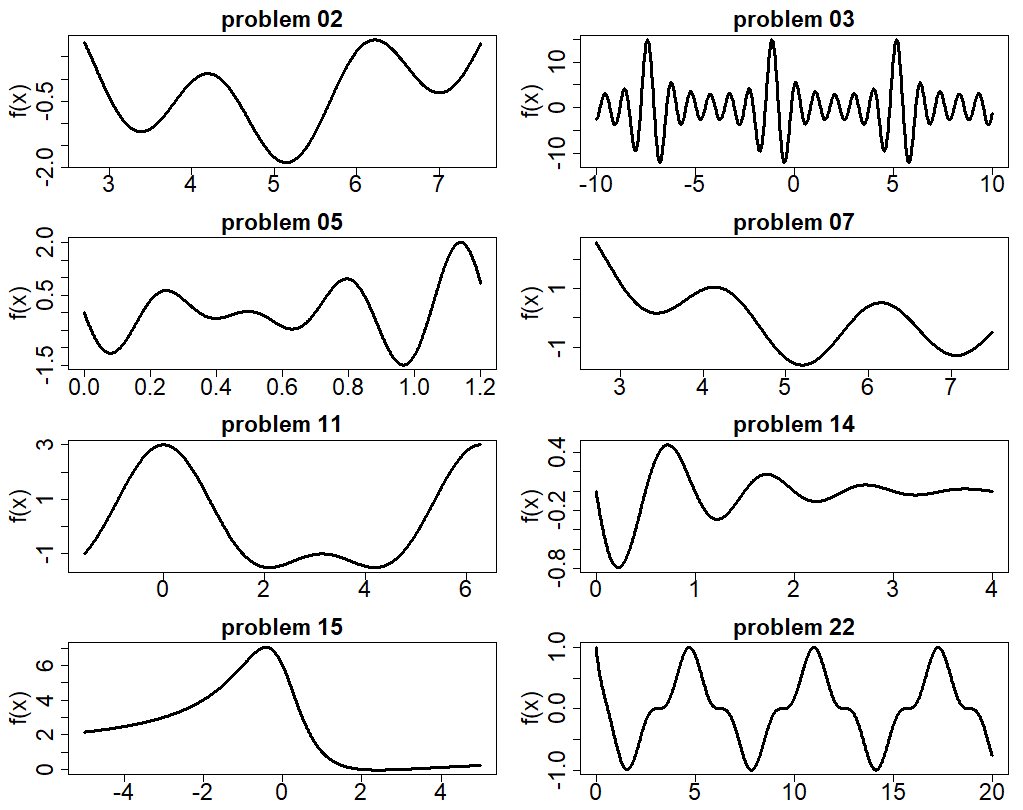}
    \caption{Graphical representations of the 1-dimensional test problems.}
    \label{fig:test_problems_1d}
\end{figure}

\begin{figure}[h!]
    \centering
    \includegraphics[width=\linewidth]{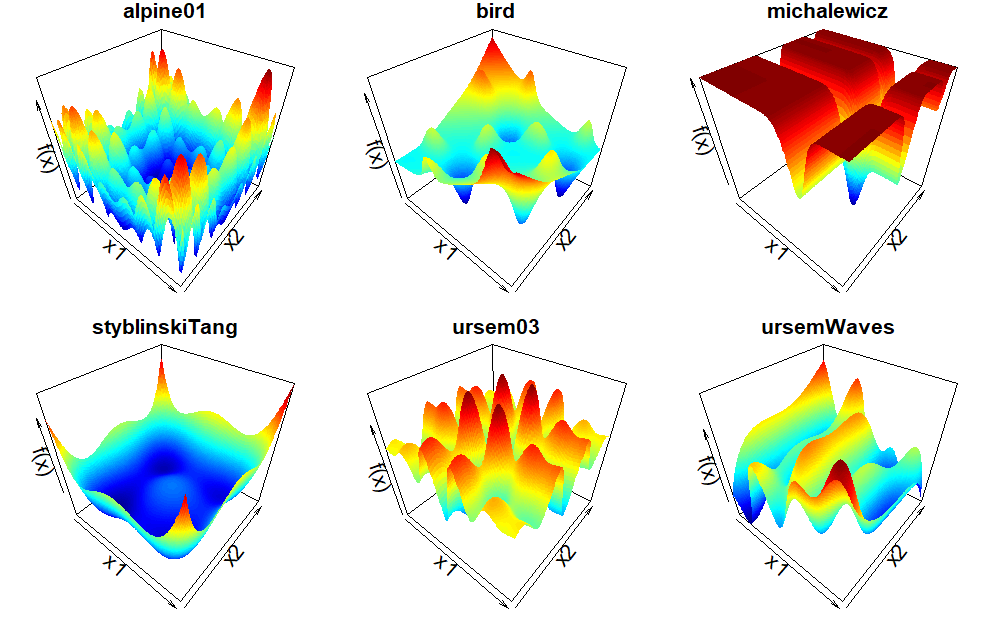}
    \caption{Graphical representations of the 2-dimensional test problems (3D view).}
    \label{fig:test_problems_2d}
\end{figure}

\begin{figure}[h!]
    \centering
    \includegraphics[width=0.95\linewidth]{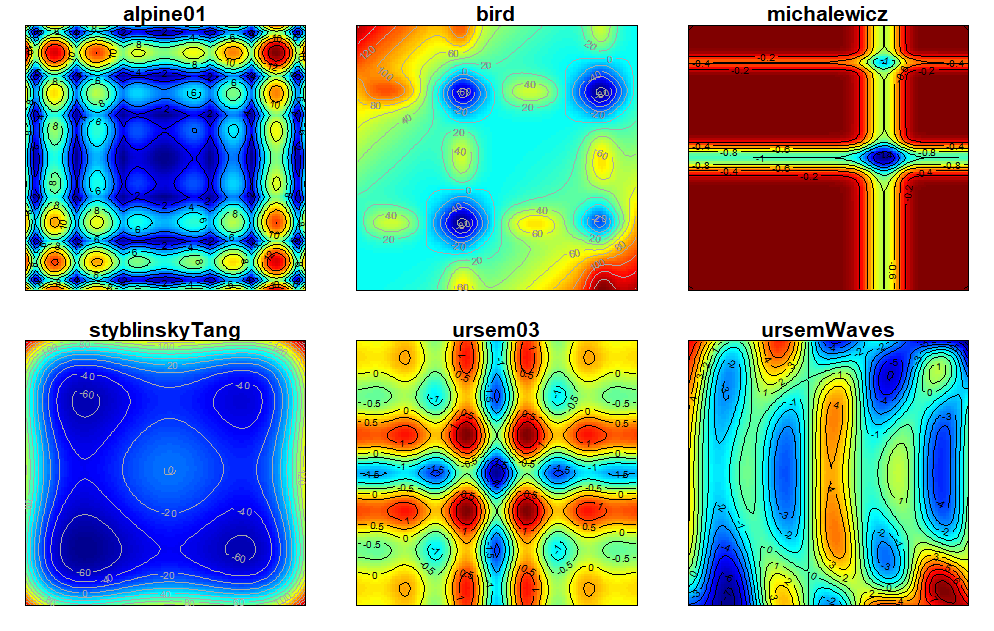}
    \caption{Graphical representations of the 2-dimensional test problems (contour plot).}
    \label{fig:test_problems_2d}
\end{figure}

\clearpage

\subsection{Test problems for multi-fidelity BO}

The multi-fidelity test problems considered in this study are taken from \cite{mainini2022analytical}, so all the equations can be retrieved from the quoted paper as well as from our github repository. Here, we summarize just the most relevant information and provide a graphical representation of the test functions.

\begin{table}[h!]
    \centering
    \resizebox{0.9\columnwidth}{!}{%
    \begin{tabular}{|l|c|c|c|}
         \hline
         Test problem & $d$ & fidelity levels & search space \\
         \hline\hline
         forrester & 1 & 4 & $[0,1]$ \\                           
         rosenbrock & any & 3 & $[-2,2]^d$ \\         
         shiftedRotatedRastrigin & any & 3 & $[-0.1,0.2]^d$ \\
         heterogenous & any & 2 & $[0,1]^d$ \\
         paciorek & any & 2 & $[0.3,1]^d$ \\         
         \hline
    \end{tabular}%
    }
    \caption{Definition of the multi-fidelity test problems addressed.}
\end{table}

\begin{figure}[h!]
    \centering
    \includegraphics[width=0.9\linewidth]{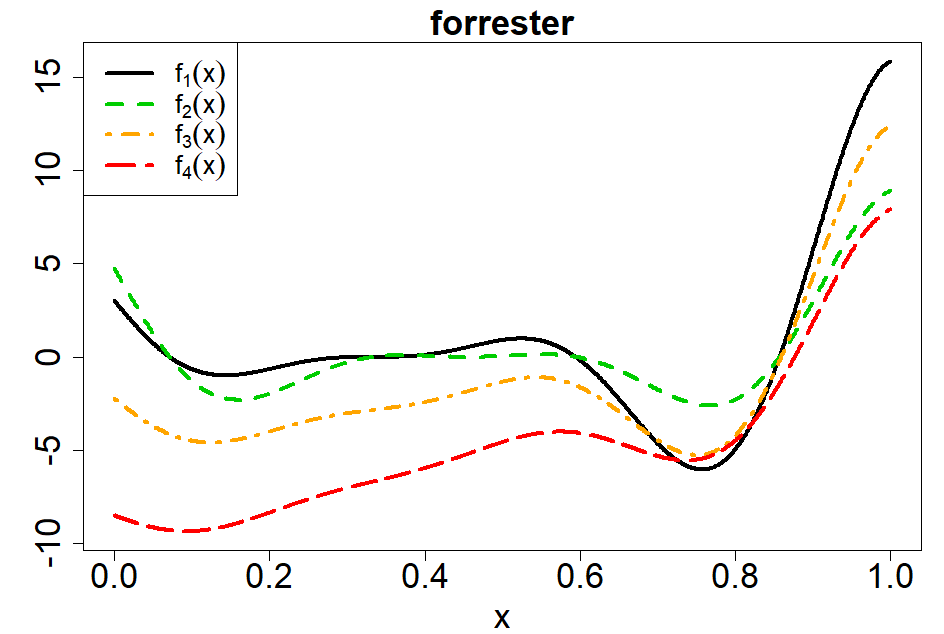}
    \caption{Graphical representation of the Forrester 1-dimensional test problem.}
    \label{fig:mf_forrester}
\end{figure}

\begin{figure}[h!]
    \centering
    \includegraphics[width=\linewidth]{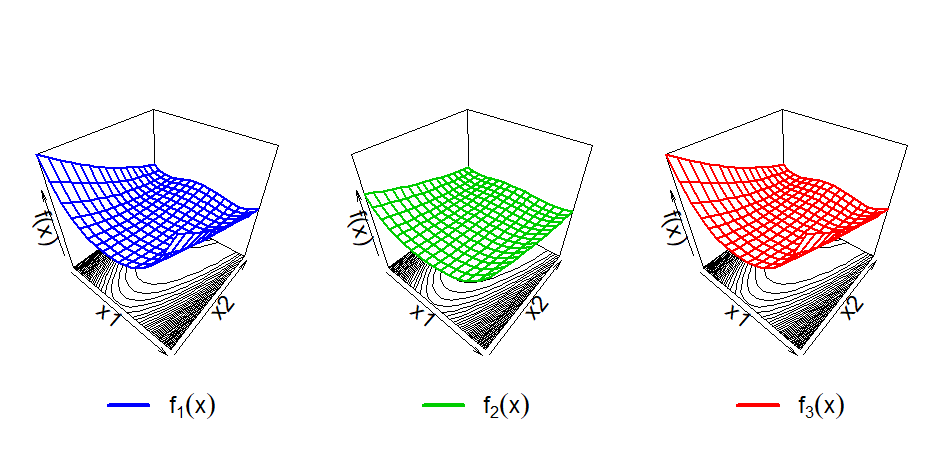}
    \caption{Graphical representation of the Rosenbrock test problem ($d=2$).}
    \label{fig:mf_rosenbrock}
\end{figure}

\begin{figure}[h!]
    \centering
    \includegraphics[width=\linewidth]{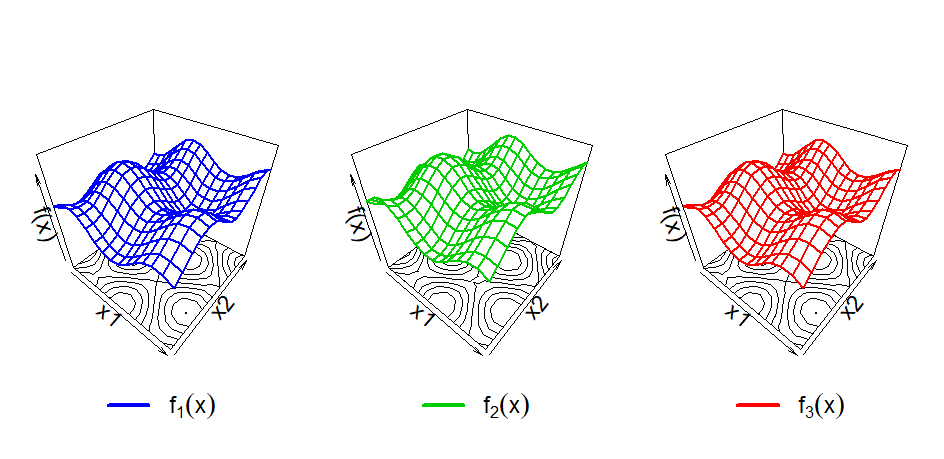}
    \caption{Graphical representation of the Shifted Rotated Rastrigin test problem ($d=2$).}
    \label{fig:mf_shiftedroratedrastrigin}
\end{figure}

\begin{figure}[h!]
    \centering
    \includegraphics[width=\linewidth]{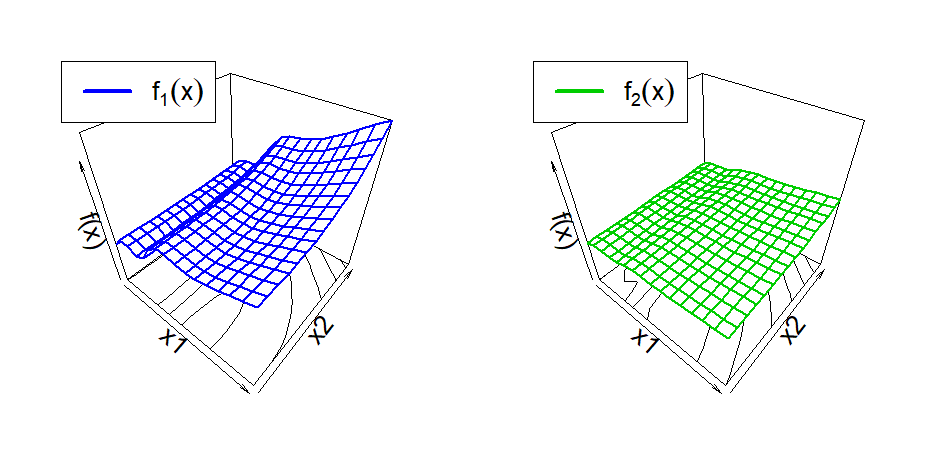}
    \caption{Graphical representation of the Heterogeneous test problem ($d=2$).}
    \label{fig:mf_heterogeneous}
\end{figure}

\begin{figure}[h!]
    \centering
    \includegraphics[width=\linewidth]{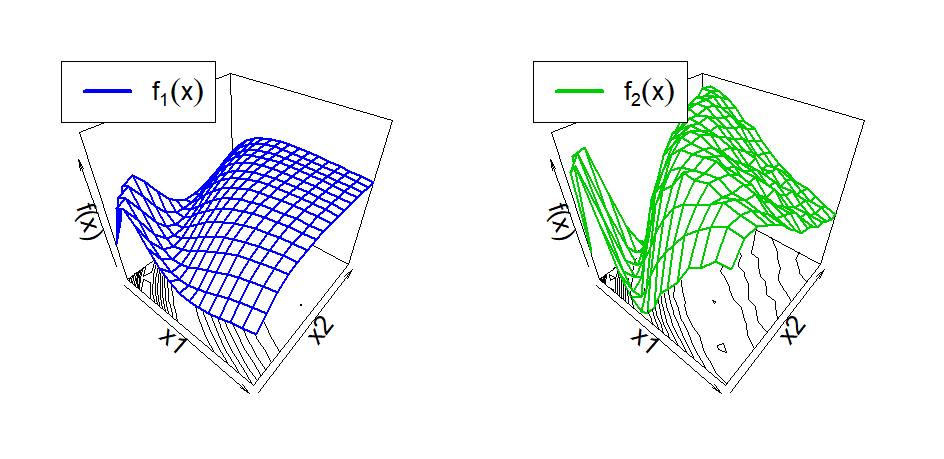}
        \caption{Graphical representation of the Pacioreck test problem ($d=2$).}
    \label{fig:mf_paciorek}
\end{figure}




\end{appendices}


\clearpage 

\bibliography{w2bgp-bo}

@article{peyre2019computational,
  title={Computational optimal transport: With applications to data science},
  author={Peyr{\'e}, Gabriel and Cuturi, Marco and others},
  journal={Foundations and Trends{\textregistered} in Machine Learning},
  volume={11},
  number={5-6},
  pages={355--607},
  year={2019},
  publisher={Now Publishers, Inc.}
}

@book{santambrogio2015optimal,
  title={Optimal transport for applied mathematicians},
  author={Santambrogio, Filippo},
  volume={87},
  year={2015},
  publisher={Springer}
}

@book{ollivier2014optimal,
  title={Optimal transport: theory and applications},
  author={Ollivier, Yann and Pajot, Herv{\'e} and Villani, C{\'e}dric},
  volume={413},
  year={2014},
  publisher={Cambridge University Press}
}

@inproceedings{arjovsky2017wasserstein,
  title={Wasserstein generative adversarial networks},
  author={Arjovsky, Martin and Chintala, Soumith and Bottou, L{\'e}on},
  booktitle={International conference on machine learning},
  pages={214--223},
  year={2017},
  organization={PMLR}
}

@article{khamis2024scalable,
  title={Scalable optimal transport methods in machine learning: A contemporary survey},
  author={Khamis, Abdelwahed and Tsuchida, Russell and Tarek, Mohamed and Rolland, Vivien and Petersson, Lars},
  journal={IEEE transactions on pattern analysis and machine intelligence},
  year={2024},
  publisher={IEEE}
}

@article{montesuma2024recent,
  title={Recent advances in optimal transport for machine learning},
  author={Montesuma, Eduardo Fernandes and Mboula, Fred Maurice Ngole and Souloumiac, Antoine},
  journal={IEEE Transactions on Pattern Analysis and Machine Intelligence},
  year={2024},
  publisher={IEEE}
}

@phdthesis{montesuma2024multi,
  title={Multi-Source Domain Adaptation through Wasserstein Barycenters},
  author={Montesuma, Eduardo Fernandes},
  year={2024},
  school={Universit{\'e} Paris-Saclay}
}

@article{gao2025wasserstein,
  title={Wasserstein convergence guarantees for a general class of score-based generative models},
  author={Gao, Xuefeng and Nguyen, Hoang M and Zhu, Lingjiong},
  journal={Journal of Machine Learning Research},
  volume={26},
  number={43},
  pages={1--54},
  year={2025}
}

@article{cheng2024convergence,
  title={Convergence of flow-based generative models via proximal gradient descent in wasserstein space},
  author={Cheng, Xiuyuan and Lu, Jianfeng and Tan, Yixin and Xie, Yao},
  journal={IEEE Transactions on Information Theory},
  year={2024},
  publisher={IEEE}
}

@article{bures1969extension,
  title={An extension of Kakutani's theorem on infinite product measures to the tensor product of semifinite w*-algebras},
  author={Bures, Donald},
  journal={Transactions of the American Mathematical Society},
  volume={135},
  pages={199--212},
  year={1969},
  publisher={JSTOR}
}

@article{forrester2016relating,
  title={Relating the Bures measure to the Cauchy two-matrix model},
  author={Forrester, Peter J and Kieburg, Mario},
  journal={Communications in Mathematical Physics},
  volume={342},
  pages={151--187},
  year={2016},
  publisher={Springer}
}

@article{agueh2011barycenters,
  title={Barycenters in the Wasserstein space},
  author={Agueh, Martial and Carlier, Guillaume},
  journal={SIAM Journal on Mathematical Analysis},
  volume={43},
  number={2},
  pages={904--924},
  year={2011},
  publisher={SIAM}
}

@book{williams2006gaussian,
  title={Gaussian processes for machine learning},
  author={Williams, Christopher KI and Rasmussen, Carl Edward},
  volume={2},
  number={3},
  year={2006},
  publisher={MIT press Cambridge, MA}
}

@book{gramacy2020surrogates,
  title={Surrogates: Gaussian process modeling, design, and optimization for the applied sciences},
  author={Gramacy, Robert B},
  year={2020},
  publisher={Chapman and Hall/CRC}
}

@article{mallasto2017learning,
  title={Learning from uncertain curves: The 2-Wasserstein metric for Gaussian processes},
  author={Mallasto, Anton and Feragen, Aasa},
  journal={Advances in Neural Information Processing Systems},
  volume={30},
  year={2017}
}

@article{wei2024scalable,
  title={Scalable Bayesian Optimization via Focalized Sparse Gaussian Processes},
  author={Wei, Yunyue and Zhuang, Vincent and Soedarmadji, Saraswati and Sui, Yanan},
  journal={arXiv preprint arXiv:2412.20375},
  year={2024}
}

@inproceedings{mcintire2016sparse,
  title={Sparse Gaussian processes for Bayesian optimization.},
  author={McIntire, Mitchell and Ratner, Daniel and Ermon, Stefano},
  booktitle={UAI},
  volume={16},
  pages={517--526},
  year={2016}
}

@article{masarotto2019procrustes,
  title={Procrustes metrics on covariance operators and optimal transportation of Gaussian processes},
  author={Masarotto, Valentina and Panaretos, Victor M and Zemel, Yoav},
  journal={Sankhya A},
  volume={81},
  number={1},
  pages={172--213},
  year={2019},
  publisher={Springer}
}

@article{hvarfner2024vanilla,
  title={Vanilla Bayesian optimization performs great in high dimensions},
  author={Hvarfner, Carl and Hellsten, Erik Orm and Nardi, Luigi},
  journal={arXiv preprint arXiv:2402.02229},
  year={2024}
}

@article{doumont2025we,
  title={We Still Don't Understand High-Dimensional Bayesian Optimization},
  author={Doumont, Colin and Fan, Donney and Maus, Natalie and Gardner, Jacob R and Moss, Henry and Pleiss, Geoff},
  journal={arXiv preprint arXiv:2512.00170},
  year={2025}
}

@article{eriksson2019scalable,
  title={Scalable global optimization via local Bayesian optimization},
  author={Eriksson, David and Pearce, Michael and Gardner, Jacob and Turner, Ryan D and Poloczek, Matthias},
  journal={Advances in neural information processing systems},
  volume={32},
  year={2019}
}

@article{zhan2025collaborative,
  title={Collaborative Bayesian Optimization via Wasserstein Barycenters},
  author={Zhan, Donglin and Zhang, Haoting and Righter, Rhonda and Zheng, Zeyu and Anderson, James},
  journal={arXiv preprint arXiv:2504.10770},
  year={2025}
}

@article{yue2025collaborative,
  title={Collaborative and Distributed Bayesian Optimization via Consensus},
  author={Yue, Xubo and Liu, Yang and Berahas, Albert S and Johnson, Blake N and Al Kontar, Raed},
  journal={IEEE Transactions on Automation Science and Engineering},
  year={2025},
  publisher={IEEE}
}

@inproceedings{al2024collaborative,
  title={Collaborative and federated black-box optimization: A Bayesian optimization perspective},
  author={Al Kontar, Raed},
  booktitle={2024 IEEE International Conference on Big Data (BigData)},
  pages={7854--7859},
  year={2024},
  organization={IEEE}
}

@inproceedings{tighineanu2022transfer,
  title={Transfer learning with gaussian processes for bayesian optimization},
  author={Tighineanu, Petru and Skubch, Kathrin and Baireuther, Paul and Reiss, Attila and Berkenkamp, Felix and Vinogradska, Julia},
  booktitle={International conference on artificial intelligence and statistics},
  pages={6152--6181},
  year={2022},
  organization={PMLR}
}

@article{feurer2018practical,
  title={Practical transfer learning for bayesian optimization},
  author={Feurer, Matthias and Letham, Benjamin and Hutter, Frank and Bakshy, Eytan},
  journal={arXiv preprint arXiv:1802.02219},
  year={2018}
}

@article{hvarfner2023self,
  title={Self-correcting bayesian optimization through bayesian active learning},
  author={Hvarfner, Carl and Hellsten, Erik and Hutter, Frank and Nardi, Luigi},
  journal={Advances in Neural Information Processing Systems},
  volume={36},
  pages={79173--79199},
  year={2023}
}

@article{candelieri2023wasserstein,
  title={Wasserstein enabled Bayesian optimization of composite functions},
  author={Candelieri, Antonio and Ponti, Andrea and Archetti, Francesco},
  journal={Journal of Ambient Intelligence and Humanized Computing},
  volume={14},
  number={8},
  pages={11263--11271},
  year={2023},
  publisher={Springer}
}

@inproceedings{candelieri2022bayesian,
  title={Bayesian optimization in Wasserstein spaces},
  author={Candelieri, Antonio and Ponti, Andrea and Archetti, Francesco},
  booktitle={International Conference on Learning and Intelligent Optimization},
  pages={248--262},
  year={2022},
  organization={Springer}
}

@article{candelieri2025bayesian,
  title={Bayesian optimization over the probability simplex},
  author={Candelieri, Antonio and Ponti, Andrea and Archetti, Francesco},
  journal={Annals of Mathematics and Artificial Intelligence},
  volume={93},
  number={1},
  pages={77--91},
  year={2025},
  publisher={Springer}
}

@article{candelieri2025multiple,
  title={Multiple Information Source Bayesian Optimization},
  author={Candelieri, Antonio and Ponti, Andrea and Archetti, Francesco},
  publisher={Springer}
}

@article{march2012provably,
  title={Provably convergent multifidelity optimization algorithm not requiring high-fidelity derivatives},
  author={March, Andrew and Willcox, Karen},
  journal={AIAA journal},
  volume={50},
  number={5},
  pages={1079--1089},
  year={2012}
}

@inproceedings{song2019general,
  title={A general framework for multi-fidelity bayesian optimization with gaussian processes},
  author={Song, Jialin and Chen, Yuxin and Yue, Yisong},
  booktitle={The 22nd International Conference on Artificial Intelligence and Statistics},
  pages={3158--3167},
  year={2019},
  organization={PMLR}
}

@inproceedings{mikkola2023multi,
  title={Multi-fidelity Bayesian optimization with unreliable information sources},
  author={Mikkola, Petrus and Martinelli, Julien and Filstroff, Louis and Kaski, Samuel},
  booktitle={International Conference on Artificial Intelligence and Statistics},
  pages={7425--7454},
  year={2023},
  organization={PMLR}
}

@article{candelieri2025wasserstein,
  title={Wasserstein Barycenter Gaussian Process based Bayesian Optimization},
  author={Candelieri, Antonio and Ponti, Andrea and Archetti, Francesco},
  journal={arXiv preprint arXiv:2505.12471},
  year={2025}
}

@inproceedings{candelieri2024mle,
  title={Mle-free gaussian process based bayesian optimization},
  author={Candelieri, Antonio and Signori, Elena},
  booktitle={International Conference on Learning and Intelligent Optimization},
  pages={81--95},
  year={2024},
  organization={Springer}
}

@article{ziomek2024time,
  title={Time-Varying Gaussian Process Bandits with Unknown Prior},
  author={Ziomek, Juliusz and Adachi, Masaki and Osborne, Michael A},
  journal={arXiv preprint arXiv:2402.01632},
  year={2024}
}

@article{candelieri2024fair,
  title={Fair and green hyperparameter optimization via multi-objective and multiple information source Bayesian optimization},
  author={Candelieri, Antonio and Ponti, Andrea and Archetti, Francesco},
  journal={Machine Learning},
  volume={113},
  number={5},
  pages={2701--2731},
  year={2024},
  publisher={Springer}
}

@book{garnett2023bayesian,
  title={Bayesian optimization},
  author={Garnett, Roman},
  year={2023},
  publisher={Cambridge University Press}
}

@book{archetti2019bayesian,
  title={Bayesian optimization and data science},
  author={Archetti, Francesco and Candelieri, Antonio},
  volume={849},
  year={2019},
  publisher={Springer}
}

@article{huang2006global,
  title={Global optimization of stochastic black-box systems via sequential kriging meta-models},
  author={Huang, Deng and Allen, Theodore T and Notz, William I and Zeng, Ning},
  journal={Journal of global optimization},
  volume={34},
  number={3},
  pages={441--466},
  year={2006},
  publisher={Springer}
}

@article{mainini2022analytical,
  title={Analytical benchmark problems for multifidelity optimization methods},
  author={Mainini, Laura and Serani, Andrea and Rumpfkeil, Markus P and Minisci, E and Quagliarella, Domenico and Pehlivan, H and Yildiz, Sihmehmet and Ficini, S and Pellegrini, Riccardo and Di Fiore, Francesco and others},
  journal={arXiv preprint arXiv:2204.07867},
  year={2022}
}

\end{document}